# Training Latent Variable Models with Auto-encoding Variational Bayes: A Tutorial

**Yang Zhi-Han**

Department of Mathematics and Statistics
Carleton College
Northfield, MN 55057
yangz2@carleton.edu

**Abstract**

**Auto-encoding Variational Bayes** (AEVB) [9] is a powerful and general algorithm for fitting latent variable models (a promising direction for unsupervised learning), and is well-known for training the Variational Auto-Encoder (VAE). In this tutorial, we focus on motivating AEVB from the classic **Expectation Maximization** (EM) algorithm, as opposed to from deterministic auto-encoders. Though natural and somewhat self-evident, the connection between EM and AEVB is not emphasized in the recent deep learning literature, and we believe that emphasizing this connection can improve the community's understanding of AEVB. In particular, we find it especially helpful to view (1) optimizing the evidence lower bound[1] (ELBO) with respect to inference parameters as **approximate E-step** and (2) optimizing ELBO with respect to generative parameters as **approximate M-step**; doing both simultaneously as in AEVB is then simply tightening and pushing up ELBO at the same time. We discuss how approximate E-step can be interpreted as performing **variational inference**. Important concepts such as amortization and the reparametrization trick are discussed in great detail. Finally, we derive from scratch the AEVB training procedures of a non-deep and several deep latent variable models, including VAE [9], Conditional VAE [19], Gaussian Mixture VAE [18] and Variational RNN [2]. It is our hope that readers would recognize AEVB as a general algorithm that can be used to fit a wide range of latent variable models (not just VAE), and apply AEVB to such models that arise in their own fields of research. PyTorch [17] code for all included models are publicly available[2].

## Table of contents

1. It is also called the variational lower bound, or the variational bound.
2. Code: https://github.com/zhihanyang2022/aevb-tutorial

# 1 Latent variable models

In probabilitic machine learning, a *model* means a (parametrized) probability distribution defined over variables of interest. This includes classifiers and regressors, which can be viewed simply as conditional distributions. A latent variable model is just a model that contains some variables whose values are not observed. Therefore, for such a model, we can divide the variables of interest into two vectors: $\boldsymbol{x}$, which denotes the vector of observed variables, and $\boldsymbol{z}$, which denotes the vector of *latent* or unobserved variables.

A strong motivation for using latent variable models is that some variables in the generative process are naturally hidden from us so we cannot observe their values. In particular, latent variables can have the interpretation of low-dimensional "hidden causes" of high-dimensional observed variables, and models that utilize latent variables "often have fewer parameters than models that directly represent correlation in the [observed] space" [15]. The low dimensionality of latent variables also means that they can serve as a compressed representation of data. Additionally, latent variable models can be highly expressive from summing over or integrating over hidden variables, which makes them useful for purposes like black-box density estimation.

## 2 Expectation maximization

In the classic statistical inference framework, fitting a model means finding the maximum likelihood estimator (MLE) of the model parameters, which is obtained by maximizing the log likelihood function (also known as the evidence[3]):

$$\boldsymbol{\theta}^* \;=\; \arg\max_{\boldsymbol{\theta}} \log p_{\boldsymbol{\theta}}(\mathcal{D}),$$

where we have assumed that the model $p_{\boldsymbol{\theta}}$ is unconditional[4] and we have $N$ i.i.d. observations of the observed variable $\boldsymbol{x}$ stored in the dataset $\mathcal{D} = \{\boldsymbol{x}_1, \boldsymbol{x}_2, \cdots, \boldsymbol{x}_N\}$.

By the same spirit, we can fit a latent variable model using MLE:

$$\begin{aligned}\boldsymbol{\theta}^* &= \arg\max_{\boldsymbol{\theta}} \log p_{\boldsymbol{\theta}}(\mathcal{D}) \\ &= \arg\max_{\boldsymbol{\theta}} \sum_{i=1}^{N} \log p_{\boldsymbol{\theta}}(\boldsymbol{x}_i) \\ &= \arg\max_{\boldsymbol{\theta}} \sum_{i=1}^{N} \log \int p_{\boldsymbol{\theta}}(\boldsymbol{x}_i, \boldsymbol{z}_i)\, d\boldsymbol{z}_i. \end{aligned} \tag{1}$$

where we have assumed that $\boldsymbol{z}_i$ is a continuous latent variable. If $\boldsymbol{z}_i$ is discrete, then the integral would be replaced by a sum. It is also valid for one part of $\boldsymbol{z}_i$ to be continuous and the other part to be discrete. In general, evaluating this integral is intractable, since it's essentially the normalization constant in Bayes' rule:

$$p_{\boldsymbol{\theta}}(\boldsymbol{z}_i |\, \boldsymbol{x}_i) = \frac{p_{\boldsymbol{\theta}}(\boldsymbol{x}_i, \boldsymbol{z}_i)}{\int p_{\boldsymbol{\theta}}(\boldsymbol{x}_i, \boldsymbol{z}_i)\, d\boldsymbol{z}_i}.$$

Note that this integral (or sum) is tractable in some simple cases, though evaluating this integral and plugging it into Equation 1 still has certain downsides (see Section 11.4.1 in [15] for a short discussion on this for Gaussian Mixture models; see [11] for a brief comment on numerical stability; it's surprisingly difficult to find sources that discuss the downsides more systematically). In other cases where it's not intractable, the reason[5] for intractability can be having no closed-form solution or computational intractability. An interested reader is encouraged to seek additional sources.

While directly optimizing the evidence is difficult, it is possible to derive a lower bound to the evidence, called the *evidence lower bound* (ELBO), as follows:

$$\begin{aligned}\sum_{i=1}^{N} \log p_{\boldsymbol{\theta}}(\boldsymbol{x}_i) &= \sum_{i=1}^{N} \log \mathbb{E}_{\boldsymbol{z}_i \sim q_i(\boldsymbol{z}_i)}\left[\frac{p_{\boldsymbol{\theta}}(\boldsymbol{x}_i, \boldsymbol{z}_i)}{q_i(\boldsymbol{z}_i)}\right] \quad \text{(introduce distributions } q_i\text{'s)} \\ &\geq \sum_{i=1}^{N} \mathbb{E}_{\boldsymbol{z}_i \sim q_i(\boldsymbol{z}_i)}\left[\log\left(\frac{p_{\boldsymbol{\theta}}(\boldsymbol{x}_i, \boldsymbol{z}_i)}{q_i(\boldsymbol{z}_i)}\right)\right] \quad \text{(apply Jensen's inequality)} \\ &= \sum_{i=1}^{N} \mathbb{E}_{\boldsymbol{z}_i \sim q_i(\boldsymbol{z}_i)}[\log p_{\boldsymbol{\theta}}(\boldsymbol{x}_i, \boldsymbol{z}_i)] + \sum_{i=1}^{N} \mathbb{H}(q_i) \\ &\triangleq \mathrm{ELBO}(\boldsymbol{\theta}, \{q_i\}),\end{aligned}$$

where the notation $\mathrm{ELBO}(\boldsymbol{\theta}, \{q_i\})$ emphasizes that ELBO is a function of $\boldsymbol{\theta}$ and $\{q_i\}$. Importantly, Jensen's inequality becomes an equality when the random variable is a constant. This happens when $q_i(\boldsymbol{z}_i) = p_{\boldsymbol{\theta}}(\boldsymbol{z}_i |\, \boldsymbol{x}_i)$ so that $p(\boldsymbol{x}_i, \boldsymbol{z}_i) / q_i(\boldsymbol{z}_i)$ becomes $p(\boldsymbol{x}_i)$, which does not contain $\boldsymbol{z}_i$. Therefore, if we keep alternating between (1) setting each $q_i(\boldsymbol{z}_i)$ to be $p(\boldsymbol{z}_i |\, \boldsymbol{x}_i)$ so that the lower bound is tight with respect to $\boldsymbol{\theta}$ and (2) maximizing the lower bound with respect to $\boldsymbol{\theta}$, then we would maximize the evidence up to a local maximum. This is known as the Expectation Maximization (EM) algorithm, and Step 1 is called the E-step and Step 2 is called the M-step. The algorithm is summarized below:

3. The name "evidence" is also commonly used for $p_{\boldsymbol{\theta}}(D)$; in this tutorial, "evidence" strictly means $\log p_{\boldsymbol{\theta}}(D)$.
4. All the derivation can be easily adapted to conditional models.
5. Someone's personal communication with David Blei: https://tinyurl.com/43auucww

**Algorithm – Expectation Maximization (EM)**

**Require.** $\mathcal{D} = \{\boldsymbol{x}_1, \dots, \boldsymbol{x}_N\}$: observed data; $\boldsymbol{\theta}_0$: initial value of parameters
$\boldsymbol{\theta} \leftarrow \boldsymbol{\theta}_0$
**while not** converged:
  $q_i(\boldsymbol{z}_i) \leftarrow p_{\boldsymbol{\theta}}(\boldsymbol{z}_i | \boldsymbol{x}_i)$ for $i = 1, \dots, N$ (Expectation step; E-step)
  $\boldsymbol{\theta} \leftarrow \arg\max_{\boldsymbol{\theta}} \text{ELBO}(\boldsymbol{\theta}, \{q_i\})$ (Maximization step: M-step)
**end while**

At this point, though we recognize that EM does correctly converge, it is not clear whether the EM approach makes things easier and what its consequences are, as compared to using Equation 1. To partly answer this question, in Section 3 and 4, we will discuss a very general and modular template for extending EM (that leads to AEVB in Section 5) to models for which the E-step (again, due to intractability of the marginalizing integral) and the M-step are not tractable. In Section 5, we will apply this general template to derive AEVB training procedures for several interesting latent variable models. Equation 1 cannot be extended in a similar fashion.

Before we move on, it's important to consider an alternative derivation (see Section 22.2.2 of [16]) of ELBO that gives us more insights on the size of the gap between the lower bound and the true objective, which is called the *variational gap*. This derivation turns out to have *great* importance for later sections. In particular, if we write the marginal $p_{\boldsymbol{\theta}}(\boldsymbol{x})$ as $p_{\boldsymbol{\theta}}(\boldsymbol{x}, \boldsymbol{z}) / p_{\boldsymbol{\theta}}(\boldsymbol{z} | \boldsymbol{x})$ instead of $\int p_{\boldsymbol{\theta}}(\boldsymbol{x}, \boldsymbol{z})\, d\boldsymbol{z}$, then we do not need to move the expectation outside the log and could have the expectation outside at the beginning. Starting from the evidence, one can show that

$$\begin{aligned}
\sum_{i=1}^{N} \log p_{\boldsymbol{\theta}}(\boldsymbol{x}_i) &= \sum_{i=1}^{N} \mathbb{E}_{\boldsymbol{z}_i \sim q_i(\boldsymbol{z}_i)}[\log p_{\boldsymbol{\theta}}(\boldsymbol{x}_i)] \\
&= \sum_{i=1}^{N} \mathbb{E}_{\boldsymbol{z}_i \sim q_i(\boldsymbol{z}_i)}\left[\log\left(\frac{p_{\boldsymbol{\theta}}(\boldsymbol{x}_i, \boldsymbol{z}_i)}{p_{\boldsymbol{\theta}}(\boldsymbol{z}_i | \boldsymbol{x}_i)}\right)\right] \\
&= \sum_{i=1}^{N} \mathbb{E}_{\boldsymbol{z}_i \sim q_i(\boldsymbol{z}_i)}\left[\log\left(\frac{p_{\boldsymbol{\theta}}(\boldsymbol{x}_i, \boldsymbol{z}_i)}{q(\boldsymbol{z}_i)} \cdot \frac{q(\boldsymbol{z}_i)}{p_{\boldsymbol{\theta}}(\boldsymbol{z}_i | \boldsymbol{x}_i)}\right)\right] \\
&= \underbrace{\sum_{i=1}^{N} \mathbb{E}_{\boldsymbol{z}_i \sim q_i(\boldsymbol{z}_i)}[\log p_{\boldsymbol{\theta}}(\boldsymbol{x}_i, \boldsymbol{z}_i) - \log q_i(\boldsymbol{z}_i)]}_{\text{ELBO}(\boldsymbol{\theta}, \{q_i\})} + \sum_{i=1}^{N} \underbrace{\mathbb{E}_{\boldsymbol{z}_i \sim q_i(\boldsymbol{z}_i)}\left[\frac{q_i(\boldsymbol{z}_i)}{p_{\boldsymbol{\theta}}(\boldsymbol{z}_i | \boldsymbol{x}_i)}\right]}_{D_{\mathbb{KL}}(q_i(\boldsymbol{z}_i) \| p_{\boldsymbol{\theta}}(\boldsymbol{z}_i | \boldsymbol{x}_i))}. \qquad (2)
\end{aligned}$$

We see that the gap between the evidence and ELBO is elegantly the sum of KL divergences between the chosen distributions $q_i$ and the true posteriors. Since $D_{\mathbb{KL}}(q_i(\boldsymbol{z}_i) \,\|\, p_{\boldsymbol{\theta}}(\boldsymbol{z}_i | \boldsymbol{x}_i)) = 0$ if and only if $q_i(\boldsymbol{z}_i) = p_{\boldsymbol{\theta}}(\boldsymbol{z}_i | \boldsymbol{x}_i)$, this agrees with our previous derivation using Jensen's inequality.

# 3 Approximate E-step as variational inference

Section 3.2 and 3.3 partly follow the treatment of [11] and [16] respectively.

## 3.1 Variational inference

To reduce the variational gap in Equation 2 (while assuming that $\boldsymbol{\theta}$ is fixed) when $p_{\boldsymbol{\theta}}(\boldsymbol{z}_i | \boldsymbol{x}_i)$ is intractable, we define a family of distributions $\mathcal{Q}$ and aim to find individual $q_i \in \mathcal{Q}$ such that the KL divergence between $q_i(\boldsymbol{z}_i)$ and $p_{\boldsymbol{\theta}}(\boldsymbol{z}_i | \boldsymbol{x}_i)$ are minimized, i.e.,

$$q_i^* = \arg\min_{q_i \in \mathcal{Q}} D_{\mathbb{KL}}(q_i(\boldsymbol{z}_i) \,\|\, p_{\boldsymbol{\theta}}(\boldsymbol{z}_i | \boldsymbol{x}_i)) \quad \text{for } i = 1, \dots, N.$$

Since this is (1) optimizing over functions (probability distributions are functions) and (2) doing inference (i.e., obtaining some representation of the true posterior), it's called "variational inference". In the Calculus of Variations, "variations" mean small changes in functions. In practice, $q_i$ would have parameters to optimize over, so we would not be directly optimizing over functions.

If the true posterior is contained in $\mathcal{Q}$ (i.e., $p_{\boldsymbol{\theta}}(\boldsymbol{z}_i | \boldsymbol{x}_i) \in \mathcal{Q}$), then clearly $q_i^* = p_{\boldsymbol{\theta}}(\boldsymbol{z}_i | \boldsymbol{x}_i)$ and setting each $q_i = q_i^*$ in Equation 2 will make ELBO a tight bound (with respect to $\boldsymbol{\theta}$) because the sum of KL divergences will be zero. Otherwise, ELBO would not be tight, but maximizing ELBO can still be useful because (1) it is by definition a lower bound to the evidence, the quantity we care about, and (2) $q_i^*$ would still be a very good approximation if $\mathcal{Q}$ is flexible so ELBO would not be too loose.

The *challenge* of this optimization problem is that the true posterior is not tractable and hence not available, so directly minimizing the KL divergence is not an option. In this case, there are two perspectives that lead to the same solution but are conceptually somewhat different:

1. From Equation 2, we see that minimizing the sum of KL divergences with respect to $q_i$'s is equivalent to maximizing ELBO with respect to $q_i$'s, since the evidence on the left-hand side is a constant with respect to $q_i$'s. Fortunately, ELBO is fairly easy to evaluate: we always know how to evaluate the unnormalized posterior $\log p_{\boldsymbol{\theta}}(\boldsymbol{x}_i, \boldsymbol{z}_i)$ for any graphical model, and the expectation operator outside can be sidestepped with techniques that we will discuss in detail in Section 5 on a per-model basis.

2. One can also minimize KL divergences with respect to $q_i$'s by dealing with the *unnormalized* true posteriors. This is a standard approach: see Section 21.2 of [15] for a textbook treatment and Section 3 of [1] for an application to deep neural networks. Importantly, the unnormalized true posterior is tractable for all graphical models, which is the foundation for techniques such as Markov Chain Monte Carlo (MCMC). Slightly abusing the notation of KL divergence (as $p_{\boldsymbol{\theta}}(\boldsymbol{x}_i, \boldsymbol{z}_i)$ is unnormalized), we have

$$\begin{aligned} D_{\mathbb{KL}}(q_i(\boldsymbol{z}_i) \,\|\, p_{\boldsymbol{\theta}}(\boldsymbol{z}_i | \boldsymbol{x}_i)) &= D_{\mathbb{KL}}(q_i(\boldsymbol{z}_i) \,\|\, p_{\boldsymbol{\theta}}(\boldsymbol{x}_i, \boldsymbol{z}_i)) + \log p_{\boldsymbol{\theta}}(\boldsymbol{x}_i) \\ &= \mathbb{E}_{\boldsymbol{z}_i \sim q_i(\boldsymbol{z}_i)}[\log q_i(\boldsymbol{z}_i) - \log p_{\boldsymbol{\theta}}(\boldsymbol{x}_i, \boldsymbol{z}_i)]. \quad \text{(dropped constant)} \end{aligned}$$

   But this is just the negation of ELBO, and minimizing the negation of ELBO with respect to $q_i$'s is equivalent to maximizing ELBO with respect to $q_i$'s.

While these two solutions are the same, they came from two different derivations and can give different insights.

## 3.2 Amortized variational inference

One natural and convenient way to define a family of distributions $\mathcal{Q}$ is through a parametrized family. However, doing this naively means that the number of parameters of $q_i$'s will grow linearly as the number of data points $\boldsymbol{x}_i$ grow. For example, if each $q_i$ is an isotropic Gaussian with parameters $\boldsymbol{\mu}_i$ and $\boldsymbol{\sigma}_i$, then $q_i$'s altogether would have $(|\boldsymbol{\mu}_i| + |\boldsymbol{\sigma}_i|) \times N$ parameters. In such cases, it is convenient to represent $q_i(\boldsymbol{z}_i)$ by a neural network $q_{\boldsymbol{\phi}}(\boldsymbol{z}_i | \boldsymbol{x}_i)$ such that $q_{\boldsymbol{\phi}}(\boldsymbol{z}_i | \boldsymbol{x}_i) = q_i(\boldsymbol{z}_i)$, where $\boldsymbol{\phi}$ is referred to as the *inference* parameters (as opposed to $\boldsymbol{\theta}$, the *generative* parameters). This approach is called *amortized* variational inference because the "cost" of having a fixed but large number of parameters gradually "pays off", in terms of memory usage and generalization benefits, as the size of the dataset grows. With amortization, the right-hand side of Equation 2 becomes

$$\underbrace{\sum_{i=1}^{N} \mathbb{E}_{\boldsymbol{z}_i \sim q_{\boldsymbol{\phi}}(\boldsymbol{z}_i|\boldsymbol{x}_i)}[\log p_{\boldsymbol{\theta}}(\boldsymbol{x}_i, \boldsymbol{z}_i) - \log q_{\boldsymbol{\phi}}(\boldsymbol{z}_i | \boldsymbol{x}_i)]}_{\mathrm{ELBO}(\boldsymbol{\theta}, \boldsymbol{\phi})} + \sum_{i=1}^{N} \underbrace{\mathbb{E}_{\boldsymbol{z}_i \sim q_{\boldsymbol{\phi}}(\boldsymbol{z}_i|\boldsymbol{x}_i)}\left[\frac{q_{\boldsymbol{\phi}}(\boldsymbol{z}_i | \boldsymbol{x}_i)}{p_{\boldsymbol{\theta}}(\boldsymbol{z}_i | \boldsymbol{x}_i)}\right]}_{D_{\mathbb{KL}}(q_{\boldsymbol{\phi}}(\boldsymbol{z}_i|\boldsymbol{x}_i) \| p_{\boldsymbol{\theta}}(\boldsymbol{z}_i|\boldsymbol{x}_i))}.$$

| *Model* | *Observed* | *Latent* | *Joint density / generative model* |
|---|---|---|---|
| FA | $\boldsymbol{x} \in \mathbb{R}^D$ | $\boldsymbol{z} \in \mathbb{R}^L$ | $\underbrace{p_{\boldsymbol{\theta}}(\boldsymbol{x} \mid \boldsymbol{z})}_{\text{Gaussian}} \underbrace{p(\boldsymbol{z})}_{\text{Gaussian}}$ |
| VAE | $\boldsymbol{x} \in \{0,1\}^{784}$ | $\boldsymbol{z} \in \mathbb{R}^L$ | $\underbrace{p_{\boldsymbol{\theta}}(\boldsymbol{x} \mid \boldsymbol{z})}_{\text{ProductOfContinuousBernoullis}} \underbrace{p(\boldsymbol{z})}_{\text{Gaussian}}$ |
| CVAE | $\boldsymbol{x} \in \{0,1\}^{784}$ | $\boldsymbol{z} \in \mathbb{R}^L$ | $\underbrace{p_{\boldsymbol{\theta}}(\boldsymbol{x} \mid \boldsymbol{z}, \boldsymbol{y})}_{\text{ProductOfContinuousBernoullis}} \underbrace{p_{\boldsymbol{\theta}}(\boldsymbol{z} \mid \boldsymbol{y})}_{\text{Gaussian}}$ |
| GMVAE | $\boldsymbol{x} \in \{0,1\}^{784}$ | $\boldsymbol{y} \in \text{OneHot}(C), \boldsymbol{z} \in \mathbb{R}^L$ | $\underbrace{p_{\boldsymbol{\theta}}(\boldsymbol{x} \mid \boldsymbol{z})}_{\text{ProductOfBernoullis}} \underbrace{p_{\boldsymbol{\theta}}(\boldsymbol{z} \mid \boldsymbol{y})}_{\text{Gaussian}} \underbrace{p(\boldsymbol{y})}_{\text{OneHotCategorical}}$ |
| VRNN | $\boldsymbol{x}_t \in \{0,1\}^{28}$ | $\boldsymbol{z}_t \in \mathbb{R}^L$ | $\prod_{t=1}^T \underbrace{p_{\boldsymbol{\theta}}(\boldsymbol{x}_t \mid \boldsymbol{x}_{<t}, \boldsymbol{z}_{\leq t})}_{\text{ProductOfBernoullis}} \underbrace{p_{\boldsymbol{\theta}}(\boldsymbol{z}_t \mid \boldsymbol{x}_{<t}, \boldsymbol{z}_t)}_{\text{Gaussian}}$ |

**Table 1.** Summary of latent variable models presented in this tutorial. FA means Factor Analysis (Section 5.2); VAE means Variational Auto-Encoder ([9]; Section 5.3); CVAE means Conditional VAE ([19]; Section 5.4); GMVAE means Gaussian Mixture VAE ([18]; Section 5.5); VRNN means Variational Recurrent Neural Network ([2]; Section 5.6). For all models except FA, the dataset used was MNIST (images of size $28 \times 28$). More specifically, we used normalized MNIST for VAE and CVAE and binarized MNIST for GMVAE and VRNN. This was done to showcase that latent variable models can have a variety of output distributions.

### 3.3 Stochastic optimization of ELBO

To minimize the size of the variational gap between ELBO and the evidence, we can maximize ELBO with respect to $\boldsymbol{\phi}$ with mini-batch gradient ascent *until convergence*:

$$\begin{aligned}\boldsymbol{\phi}^{t+1} &\leftarrow \boldsymbol{\phi}^t + \eta \nabla_{\boldsymbol{\phi}} \Bigg\{ \underbrace{\frac{1}{N_B} \sum_{i=1}^{N_B} \underbrace{\mathbb{E}_{\boldsymbol{z}_i \sim q_{\boldsymbol{\phi}}(\boldsymbol{z}_i \mid \boldsymbol{x}_i)}[\log p_{\boldsymbol{\theta}}(\boldsymbol{x}_i, \boldsymbol{z}_i) - \log q_{\boldsymbol{\phi}}(\boldsymbol{z}_i \mid \boldsymbol{x}_i)]}_{\text{per-example ELBO}}}_{\text{mini-batch ELBO}} \Bigg\}_{\boldsymbol{\phi}=\boldsymbol{\phi}_t} \\ &= \boldsymbol{\phi}^t + \eta \frac{1}{N_B} \sum_{i=1}^{N_B} \nabla_{\boldsymbol{\phi}} \{\mathbb{E}_{\boldsymbol{z}_i \sim q_{\boldsymbol{\phi}}(\boldsymbol{z}_i \mid \boldsymbol{x}_i)}[\log p_{\boldsymbol{\theta}}(\boldsymbol{x}_i, \boldsymbol{z}_i) - \log q_{\boldsymbol{\phi}}(\boldsymbol{z}_i \mid \boldsymbol{x}_i)]\}_{\boldsymbol{\phi}=\boldsymbol{\phi}_t},\end{aligned} \tag{3}$$

where $\eta > 0$ is the learning rate and $N_B$ is the batch size. We have divided the mini-batch ELBO by $N_B$ so that picking $\eta$ can be de-coupled from picking $N_B$. However, unless the expectations within the gradient operators are tractable, the gradients cannot be evaluated exactly. We will discuss solutions to this problem on a per-model basis in Section 5.

# 4 Approximate M-step

After an approximate E-step is completed, we can then maximize ELBO with respect to $\boldsymbol{\theta}$ with mini-batch gradient ascent *until convergence*:

$$\begin{aligned}\boldsymbol{\theta}^{t+1} &\leftarrow \boldsymbol{\theta}^t + \eta \nabla_{\boldsymbol{\theta}} \Bigg\{ \frac{1}{N_B} \sum_{i=1}^{N_B} \mathbb{E}_{\boldsymbol{z}_i \sim q_{\boldsymbol{\phi}}(\boldsymbol{z}_i \mid \boldsymbol{x}_i)}[\log p_{\boldsymbol{\theta}}(\boldsymbol{x}_i, \boldsymbol{z}_i) - \log q_{\boldsymbol{\phi}}(\boldsymbol{z}_i \mid \boldsymbol{x}_i)] \Bigg\}_{\boldsymbol{\theta}=\boldsymbol{\theta}_t} \\ &= \boldsymbol{\theta}^t + \eta \frac{1}{N_B} \sum_{i=1}^{N_B} \nabla_{\boldsymbol{\theta}} \{\mathbb{E}_{\boldsymbol{z}_i \sim q_{\boldsymbol{\phi}}(\boldsymbol{z}_i \mid \boldsymbol{x}_i)}[\log p_{\boldsymbol{\theta}}(\boldsymbol{x}_i, \boldsymbol{z}_i) - \log q_{\boldsymbol{\phi}}(\boldsymbol{z}_i \mid \boldsymbol{x}_i)]\}_{\boldsymbol{\theta}=\boldsymbol{\theta}_t}.\end{aligned} \tag{4}$$

Unlike in approximate E-step, here the gradient operator with respect to $\boldsymbol{\theta}$ can be moved *inside* the expectation and the expectation can be sampled. In Section 5, we will discuss this in more detail.

# 5 Derivation of AEVB for a few latent variable models

In this section, we will derive the AEVB training procedure for models listed in Table 1.

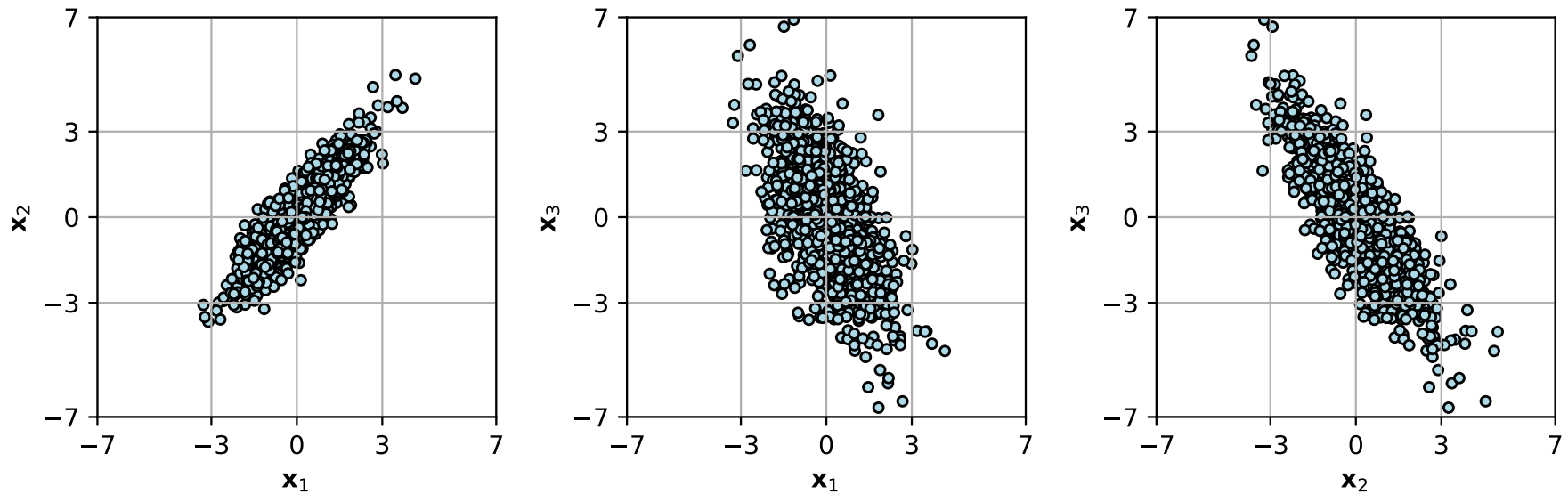

**Figure 1.** Synthetic dataset ($n = 1000$) generated by a factor analysis model with $L = 2$ and $D = 3$.

## 5.1 What exactly is the AEVB algorithm?

So far, we have discussed how E-step and M-step in EM can be approximated. However, we haven't yet arrived at AEVB[6]. Compared to just performing approximate E-steps and M-steps, AEVB makes the following additional changes. Firstly, instead of waiting for approximate E-step and M-step to converge before moving onto one another, AEVB performs gradient ascent with respect to $\boldsymbol{\phi}, \boldsymbol{\theta}$ simultaneously. This can have the advantage of fast convergence, as we share see Section 5.2.4. Secondly, the "AE" part refers to using a specific unbiased, low-variance and easy-to-evalaute estimator for the per-example ELBO such that the gradient of that estimator (with respect $\boldsymbol{\phi}$ and $\boldsymbol{\theta}$) is an unbiased estimator of the gradient of the per-example ELBO. In this section, we will derive this estimator for several interesting models and showcase PyTorch code snippets for implementing the generative, inferential and algorithmic components.

## 5.2 Factor analysis model

### 5.2.1 Generative model

The factor analysis (FA) model is the generative model defined as follows:

$$\begin{aligned} \boldsymbol{x}_i &\sim \mathcal{N}(\boldsymbol{W}\boldsymbol{z}_i, \boldsymbol{\Phi}) \\ \boldsymbol{z}_i &\sim \mathcal{N}(0, \boldsymbol{I}_L) \end{aligned}$$

where $\boldsymbol{z}_i \in \mathbb{R}^L$ is the latent variable, $\boldsymbol{x}_i \in \mathbb{R}^D$ is the observed variable, $\boldsymbol{W} \in \mathbb{R}^{D \times L}$ is the *factor loading* matrix and $\boldsymbol{\Phi}$ is a diagonal covariance matrix. The observed variable $\boldsymbol{x}$ is "generated" by linearly transforming the latent variable $\boldsymbol{z}$ and adding diagonal gaussian noise. We have assumed that $\boldsymbol{x}$ has zero mean, since it's trivial to de-mean a dataset.

For simplicity, we fit a low-dimensional FA model with $L = 2$ and $D = 3$ to a synthetic dataset generated by a ground-truth FA model with $L = 2$ and $D = 3$. Due to the difficulty of visualizing 3-dimensional data, we show the data projected $x_1$-$x_2$, $x_1$-$x_3$ and $x_2$-$x_3$ planes (Figure 1). The goal is to see whether AEVB can be used to successfully fit the FA model.

The FA model to be fitted can be defined as a PyTorch[7] module, which conveniently allows for learnable $\boldsymbol{W}$ and learnable standard deviation vector (which is the diagonal of $\boldsymbol{\Phi}$):

```
class p_x_given_z_class(nn.Module):

    def __init__(self):
        super().__init__()
        self.W = nn.Parameter(data=torch.randn(3, 2))
```

6. We have already gotten to the "VB" part; the "VB" part refers to the fact that the approximate E-step is essentially (amortized) variational inference, which is also commonly referred to as the Variational Bayesian approach.

7. `nn` is the short-hand for `torch.nn`; `Ind` is the short-hand for `torch.distributions.Independent`; `Normal` is the short-hand for `torch.distributions.Normal`.

```
        self.pre_sigma = nn.Parameter(data=torch.randn(3))

    @property
    def sigma(self):
        return F.softplus(self.pre_sigma)

    def forward(self, zs):
        # zs shape: (batch size, 2)
        mus = (self.W @ zs.T).T  # mus shape: (batch size, 3)
        return Ind(Normal(mus, sigma), reinterpreted_batch_ndims=1)  # sigma shape: (3, )
```

### 5.2.2 Approximate posterior

Running AEVB requires that we define a family of approximate posteriors $q_{\boldsymbol{\phi}}(\boldsymbol{z}_i | \boldsymbol{x}_i)$. Fortunately, for an FA model, analytic results are available (see 12.1.2 of [15]): one can show that the exact posterior $p_{\boldsymbol{\theta}}(\boldsymbol{z}_i | \boldsymbol{x}_i)$ is a Gaussian whose mean is related to $\boldsymbol{x}$ by a linear transformation (we shall denote this matrix by $\boldsymbol{V}$) and whose covariance matrix is full but independent of $\boldsymbol{x}$ (we shall denote this matrix by $\boldsymbol{\Sigma}$). Therefore, we can simply pick such Gaussians as the parametrized family[8] ($\boldsymbol{\phi} = (\boldsymbol{V}, \boldsymbol{\Sigma})$), and define its PyTorch[9] module:

```
class q_z_given_x_class(nn.Module):

    def __init__(self):
        super().__init__()
        self.V = nn.Parameter(data=torch.randn(2, 3))
        self.cov_decomp = nn.Parameter(torch.cholesky(torch.eye(2), upper=True))

    @property
    def cov(self):
        temp = torch.triu(self.cov_decomp)
        return temp.T @ temp

    def forward(self, xs):
        # xs shape: (batch size, 3)
        mus = (self.V @ xs.T).T
        return MNormal(mus, self.cov)
```

where we have followed the standard practice to learn the *Cholesky decomposition* of the covariance matrix rather than the covariance matrix directly, since entries of the decomposition are unconstrained (i.e., real numbers) and hence more amendable to gradient-based optimization.

### 5.2.3 Estimator of per-example ELBO

Recall that, in both the E-step and the M-step, the primary challenge is that we need to compute gradients of per-example ELBOs: since the expectation operators contained in per-example ELBOs are not assumed to be tractable, we cannot evaluate these gradients exactly and must resort to using estimators for these gradients. It turns out that we can first construct an unbiased, low-variance estimator for each expectation rather than for each gradient; then, as long as the source of variability of this estimator does *not* depend on the parameters (which is already true for $\boldsymbol{\theta}$), the gradient of this unbiased estimator would be an unbiased estimator of the gradient. To achieve this "independence", we apply the *reparametrization trick* to the per-example ELBO as follows:

$$
\begin{aligned}
& \mathbb{E}_{\boldsymbol{z}_i \sim q_{\boldsymbol{\phi}}(\boldsymbol{z}_i | \boldsymbol{x}_i)}[\log p_{\boldsymbol{\theta}}(\boldsymbol{x}_i, \boldsymbol{z}_i) - \log q_{\boldsymbol{\phi}}(\boldsymbol{z}_i | \boldsymbol{x}_i)] \\
= \; & \mathbb{E}_{\boldsymbol{z}_i \sim q_{\boldsymbol{\phi}}(\boldsymbol{z}_i | \boldsymbol{x}_i)}[\log p_{\boldsymbol{\theta}}(\boldsymbol{x}_i | \boldsymbol{z}_i) + \log p(\boldsymbol{z}_i) - \log q_{\boldsymbol{\phi}}(\boldsymbol{z}_i | \boldsymbol{x}_i)] \\
= \; & \mathbb{E}_{\boldsymbol{z}_i \sim q_{\boldsymbol{\phi}}(\boldsymbol{z}_i | \boldsymbol{x}_i)}[\log p_{\boldsymbol{\theta}}(\boldsymbol{x}_i | \boldsymbol{z}_i)] - D_{\mathbb{KL}}(q_{\boldsymbol{\phi}}(\boldsymbol{z}_i | \boldsymbol{x}_i) \, \| \, p(\boldsymbol{z}_i)) \\
= \; & \mathbb{E}_{\underbrace{\varepsilon_i \sim \mathcal{N}(0, \boldsymbol{I}_2)}_{\text{No longer involves } \boldsymbol{\phi}!}}[\log p_{\boldsymbol{\theta}}(\boldsymbol{x}_i | \boldsymbol{z}_i^s)] - D_{\mathbb{KL}}(q_{\boldsymbol{\phi}}(\boldsymbol{z}_i | \boldsymbol{x}_i) \, \| \, p(\boldsymbol{z}_i)) \quad (\boldsymbol{z}_i^s = \boldsymbol{V}\boldsymbol{x}_i + \text{cholesky}(\boldsymbol{\Sigma})\varepsilon_i),
\end{aligned}
$$

8. This family contains the true posterior, so doing variational inference for FA turns out to be doing exact inference, except that we don't need to derive the complicated closed-form formulas for $\boldsymbol{V}$ and $\boldsymbol{\Sigma}$.

9. `MNormal` is the short-hand for `torch.distributions.MultivariateNormal`.

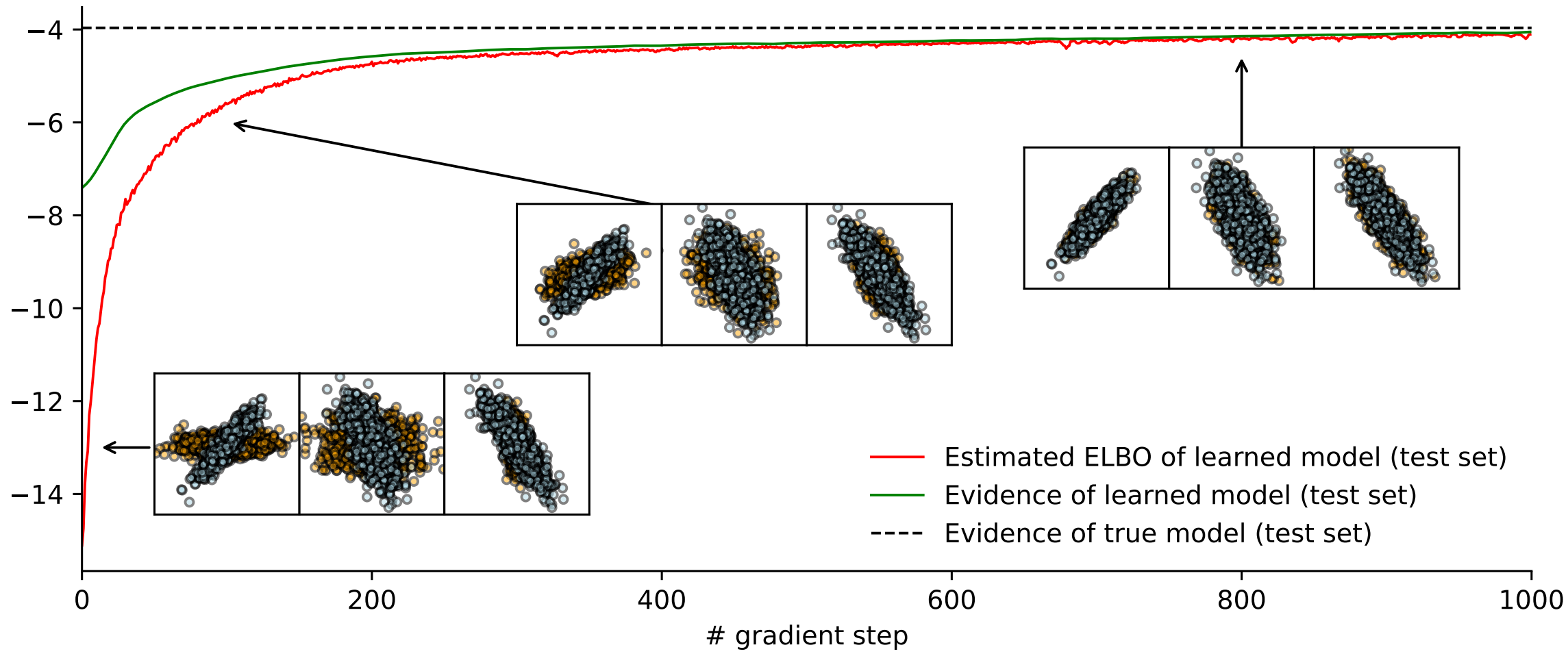

**Figure 2.** Test set performance of the FA model across training. Red and green curves show that estimated ELBO and evidence improves towards the evidence of the true model (black dotted line) respectively, and that ELBO is indeed a lower bound to the evidence. As ELBO improves, we see that generated data (orange points) gradually matches test data (blue points) in distribution.

where the second KL term can be evaluated in closed form for Gaussians (which is the case for FA), and the source of randomness in the first expectation is indeed no longer depend on $\boldsymbol{\phi}$. We have used $\boldsymbol{z}_i^s$ to denote the *reparametrized sample*. This reparametrized expression allows us to instantiate the following unbiased, low-variance estimator of the per-example ELBO:

$$\widehat{\mathrm{ELBO}}(\boldsymbol{x}_i, \boldsymbol{\theta}, \boldsymbol{\phi}) = \log p_{\boldsymbol{\theta}}(\boldsymbol{x}_i | \boldsymbol{z}_i^s) - D_{\mathbb{KL}}(q_{\boldsymbol{\phi}}(\boldsymbol{z}_i | \boldsymbol{x}_i) \,\|\, p_{\boldsymbol{\theta}}(\boldsymbol{z}_i)) \quad (\boldsymbol{z}_i^s = \boldsymbol{V}\boldsymbol{x}_i + \mathrm{cholesky}(\boldsymbol{\Sigma})\boldsymbol{\varepsilon}_i),$$

where $\log p_{\boldsymbol{\theta}}(\boldsymbol{x}_i | \boldsymbol{z}_i^s)$ intuitively measures how well the generative and inference components collaboratively perform *reconstruction* or *auto-encoding*[10]: encoding $\boldsymbol{x}_i$ probabilistically into $\boldsymbol{z}_i^s$, and then decoding $\boldsymbol{z}_i$ deterministically into $\boldsymbol{x}_i^s$. In PyTorch[11], we can implement this estimator and compute the gradient of it with respect to $\boldsymbol{\theta}$ and $\boldsymbol{\phi}$ as follows:

```
class AEVB(nn.Module):
    # ...
    def step(self, xs):
        # xs shape: (batch size, 3)
        posterior_over_zs = self.q_z_given_x(xs)
        kl = D.kl.kl_divergence(posterior_over_zs, self.p_z)
        zs = posterior_over_zs.rsample()  # reparametrized samples
        rec = self.p_x_given_z(zs).log_prob(xs)  # reconstruction
        per_example_elbos = rec - kl  # values of estimators of per-example ELBOs
        mini_batch_elbo = per_example_elbos.mean()
        loss = - mini_batch_elbo
        self.optimizer.zero_grad()
        loss.backward()
        self.optimizer.step()
    # ...
```

### 5.2.4 Results

**Experiment 1.** In this experiment, we empirically assess the convergence of AEVB on our simple FA model. We run gradient ascent on estimated mini-batch ELBO[12], and after every gradient step we measure performance of AEVB on two metrics: the estimated ELBO on the entire test set ($n = 1000$) and the exact evidence on the entire test set. For training hyperparameters, we used a mini-batch size of 32 and Adam [8] with a learning rate of 1e-2.

10. Hence the name *Auto-Encoding* Variational Bayes. If the expectation within the KL term is also sampled (which yields a higher-variance estimator), the algorithm is instead called *Stochastic Gradient* Variational Bayes (SGVB).

11. `kl_divergence` is the short-hand for `torch.distributions.kl.kl_divergence`. Also, note that we used `rsample` instead of `sample` – this is *crucial*; otherwise we would not be able to differentiate through $\boldsymbol{z}_i^s$.

12. This is simply an average of estimators of per-example ELBOs, as shown in the code snippet.

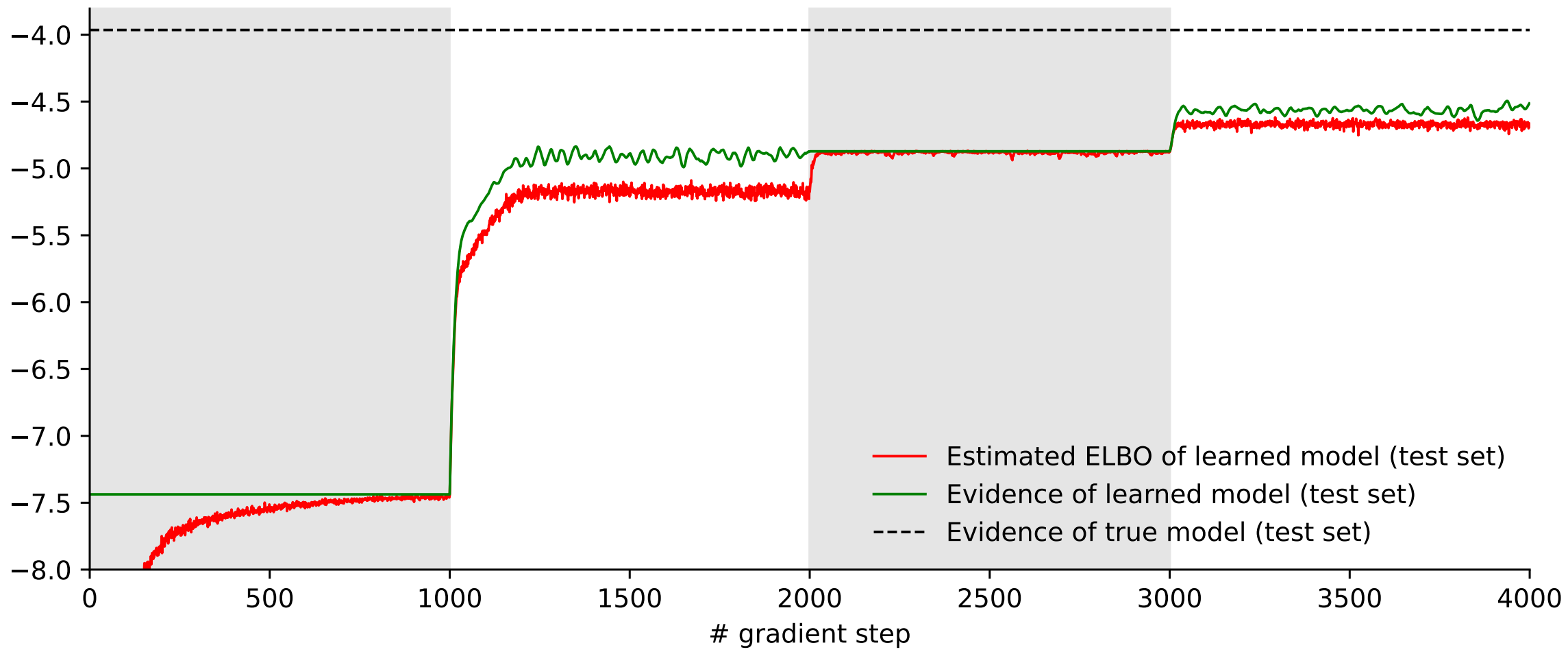

**Figure 3.** Test set performance of the FA model across training when alternating between periods of only updating inference parameters $\boldsymbol{\phi}$ (gray regions) and periods of only updating generative parameters $\boldsymbol{\theta}$ (white regions). In gray regions, ELBO becomes tight; in white regions, both ELBO and evidence improves but ELBO is no longer tight.

It is worth noting that, for arbitrary latent variable models, the evidence is generally intractable since it involves accurately evaluating the integral in Equation 1. While Monte Carlo integration (by first obtaining samples from $p(\boldsymbol{z}_i)$ and then averaging $p(\boldsymbol{x}_i | \boldsymbol{z}_i)$) can work, it has high variance when the latent space has high dimensionality and requires too much computation to process all the samples. Fortunately, the evidence can be expressed analytically for an FA model:

```
class AEVB(nn.Module):
    # ...
    def compute_evidence(self, xs):
        # xs shape: (batch size, 3)
        W = self.p_x_given_z.W
        Phi = (self.p_x_given_z.sigma * torch.eye(3)) ** 2
        p_x = MNormal(torch.zeros(3), Phi + W @ W.T)  # Phi + W @ W.T is low-rank approximation
        return float(p_x.log_prob(xs).mean())
    # ...
```

During training, we also generate data from the learned FA model to check whether they match the test data in distribution. This check is similar to a posterior predictive check in Bayesian model fitting, which allows us to see whether the model of choice is appropriate (e.g., does it underfit?). Finally, we note that the learned FA model will not recover the parameters of the true FA model, since the matrix $\boldsymbol{W}$ is only unique up to a right-hand side multiplication with a 2x2 rotation matrix.

Figure 2 shows the results of this experiment. We see that estimated ELBO and evidence on the test set improve over time, eventually coming very close to the evidence of the true model. In particular, ELBO is indeed a lower bound to the evidence. Predictive checks at different stages of training shows that, as ELBO improves, generated data is closer and closer to and eventually indistinguishable from test data . These results suggest that AEVB was successful at fitting the FA model. Empirically, we also find less expressive approximate posteriors (e.g., with a diagonal covariance matrix, or with a diagonal covariance matrix with fixed[13] entries) to work reasonaly well.

**Experiment 2.** In this experiment, we confirm that viewing AEVB through the lense of EM is a reasonable one. In particular, we verify the following: if we keep updating the inference parameters without updating the generative parameters (approximate E-step), then ELBO would become a tight lower bound to the evidence; if we then switch to keep updating the generative parameters without updating the inference parameters (approximate M-step), then both ELBO and evidence would improve but ELBO would no longer be a tight lower bound. Hyperparameter values are unchanged from Experiment 1.

13. This works well only when the fixed values are small.

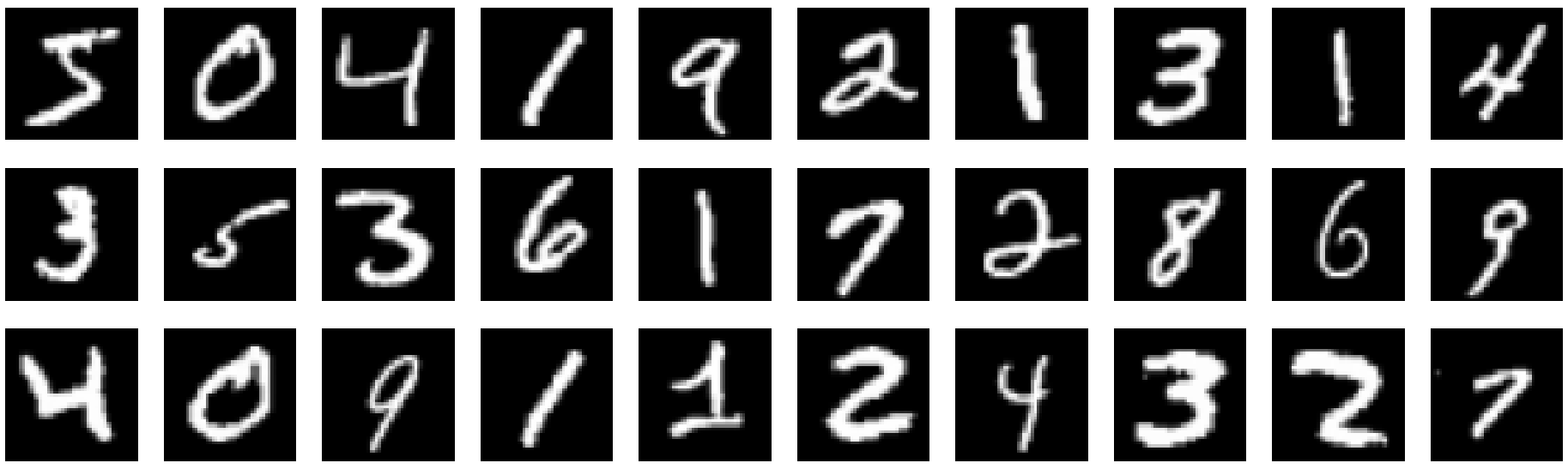

**Figure 4.** Original MNIST images.

The experiment is designed as follows. We alternate between only updating inference parameters for 1000 gradient steps and only updating generative parameters for 1000 gradient steps for two rounds, which adds up 4000 gradient steps in total. As in Experiment 1, we are tracking the estimated ELBO and evidence evaluated on the entire test set after gradient step.

Figure 3 shows the results for this experiment. Gray regions represent periods during which we only updated the inference aprameters, and white regions represent periods during which we only updated the generative parameters. In gray regions, we see that the evidence is fixed while ELBO gradually becomes a tight lower bound. From hindsight, this shouldn't be surprising, since inference parameters do not participate in the evidence computation. In white regions, we see that both ELBO and evidence improves, but ELBO is no longer a tight lower bound.

## 5.3 Variational autoencoder

### 5.3.1 Generative model

The generative part of VAE [9] for normalized[14] MNIST images (Figure 4) is defined as follows:

$$\begin{aligned} \boldsymbol{x}_i &\sim \text{Product-Of-Continuous-Bernoullis}(\boldsymbol{\lambda}_{\boldsymbol{\theta}}(\boldsymbol{z}_i)) \\ \boldsymbol{z}_i &\sim \mathcal{N}(0, \boldsymbol{I}_L) \end{aligned}$$

where $\boldsymbol{z}_i \in \mathbb{R}^L$ is the latent variable and $\boldsymbol{x}_i \in \mathbb{R}^D$ is the observed variable. In particular, $D = 28 \times 28$ where 28 is the height and width of each image. Product-Of-Continuous-Bernoullis is a product of $D$ independent continuous Bernoulli distributions ([12]; each of which has support $[0, 1]$ with parameter $\lambda \in [0, 1]$) with PDF

$$p(\boldsymbol{x}_i | \boldsymbol{z}_i) = \prod_{j=1}^{D} \text{Continuous-Bernoulli}(x_{ij} | \lambda_{ij}) \quad \text{where} \quad \lambda_{ij} = \boldsymbol{\lambda}_{\boldsymbol{\theta}}(\boldsymbol{z}_i)_j$$

where $\boldsymbol{\lambda}_{\boldsymbol{\theta}}: \mathbb{R}^L \rightarrow [0, 1]^D$ is a neural network that maps latent vectors to parameter vectors of the product of independent continuous Bernoullis. We can define this generative model in PyTorch[15] (following the network architecture in the official TensorFlow code[16] for [12]):

```
class p_x_given_z_class(nn.Module):
```

14. Original MNIST is on the scale of 0-255 (discrete). We follow the common practice of adding Uniform noise from 0 to 1 to each pixel and then dividing by 256 for normalization.

15. `CB` is the short-hand for `torch.distributions.ContinuousBernoulli`.

16. Code: `https://github.com/cunningham-lab/cb_and_cc/blob/master/cb/utils.py`

```python
    def __init__(self, z_dim=20, x_dim=28*28):
        super().__init__()
        self.z_dim = z_dim
        self.x_dim = x_dim
        self.lambdas = nn.Sequential(
            nn.Linear(self.z_dim, 500),
            nn.ReLU(),
            nn.Dropout(0.1),
            nn.Linear(500, 500),
            nn.ReLU(),
            nn.Dropout(0.1),
            nn.Linear(500, self.x_dim),
            nn.Sigmoid()
        )

    def forward(self, zs):
        # zs shape: (batch size, L)
        return Ind(CB(self.lambdas(zs)), 1)
```

Of course, the choice of output distribution is due to the nature of normalized MNIST data that we are trying to model[17], and other choices are needed for other types of data. For example, for binarized MNIST, Bernoulli distributions would be sufficient. Also, our choice is not perfect, since it does not capture the correlation of neighboring pixels, which has consequences that we will discuss.

### 5.3.2 Approximate posterior

Unlike for FA models, the posterior for a VAE model is not tractable, since the relationship between the latent and observed variables, as specified by the two neural networks $\boldsymbol{\mu}_{\boldsymbol{\theta}}$ and $\boldsymbol{\sigma}_{\boldsymbol{\theta}}$, is highly nonlinear. To define a flexible enough family of approximate posterior to represent the true posterior with enough accuracy, we also use a neural network, though in practice the output distribution can be relatively simple, e.g., a diagonal Gaussian:

$$q_{\boldsymbol{\phi}}(\boldsymbol{z}_i | \boldsymbol{x}_i) = \mathcal{N}(\boldsymbol{\mu}_{\boldsymbol{\phi}}(\boldsymbol{x}_i), \boldsymbol{\sigma}_{\boldsymbol{\phi}}(\boldsymbol{x}_i))$$

which can be implemented as (again, following official code of [12]):

```python
class q_z_given_x_class(nn.Module):

    def __init__(self, z_dim, x_dim):
        super().__init__()
        self.z_dim = z_dim
        self.x_dim = x_dim
        self.shared = nn.Sequential(
            nn.Linear(x_dim, 500),
            nn.ReLU(),
            nn.Dropout(0.1),
            nn.Linear(500, 500),
            nn.ReLU(),
            nn.Dropout(0.1),
        )
        self.mus = nn.Linear(500, z_dim)
        self.sigmas = nn.Sequential(
            nn.Linear(500, z_dim),
            nn.Softplus()
        )

    def forward(self, xs):
        # xs shape: (batch size, 28 * 28)
```

17. In fact, any product of distributions defined over the interval $[0, 1]$ would be reasonable, e.g., the Beta distribution.

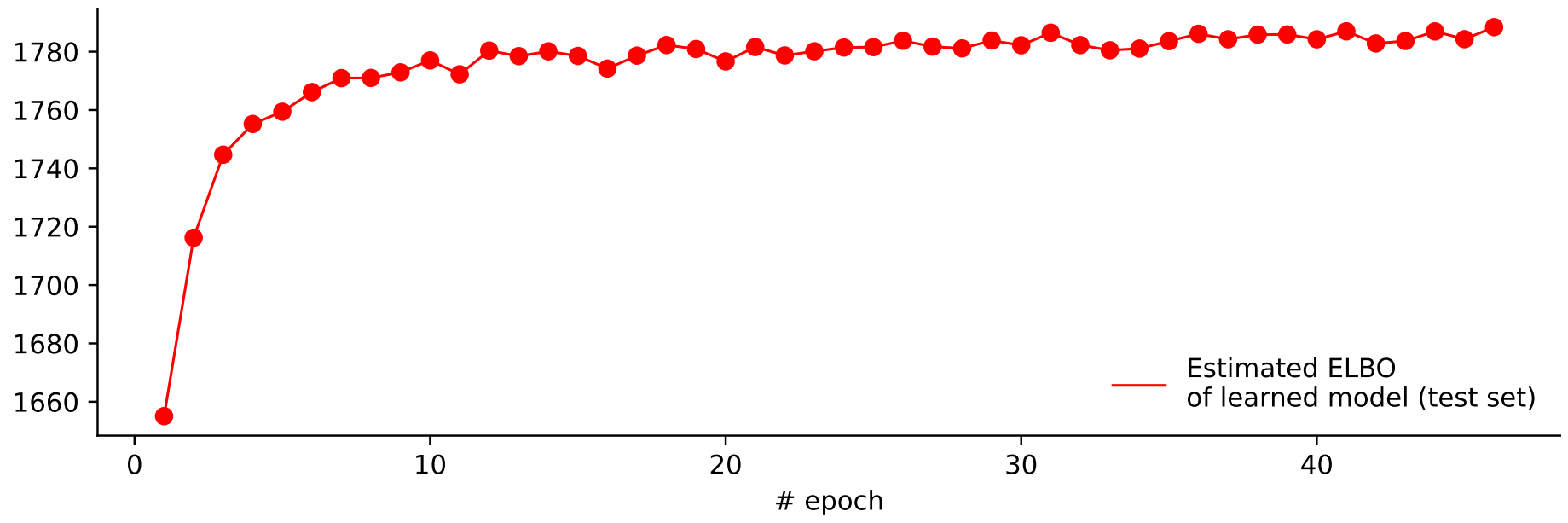

**Figure 5.** Estimated ELBO (test set) of VAE across training. It first improves quickly then plateaus.

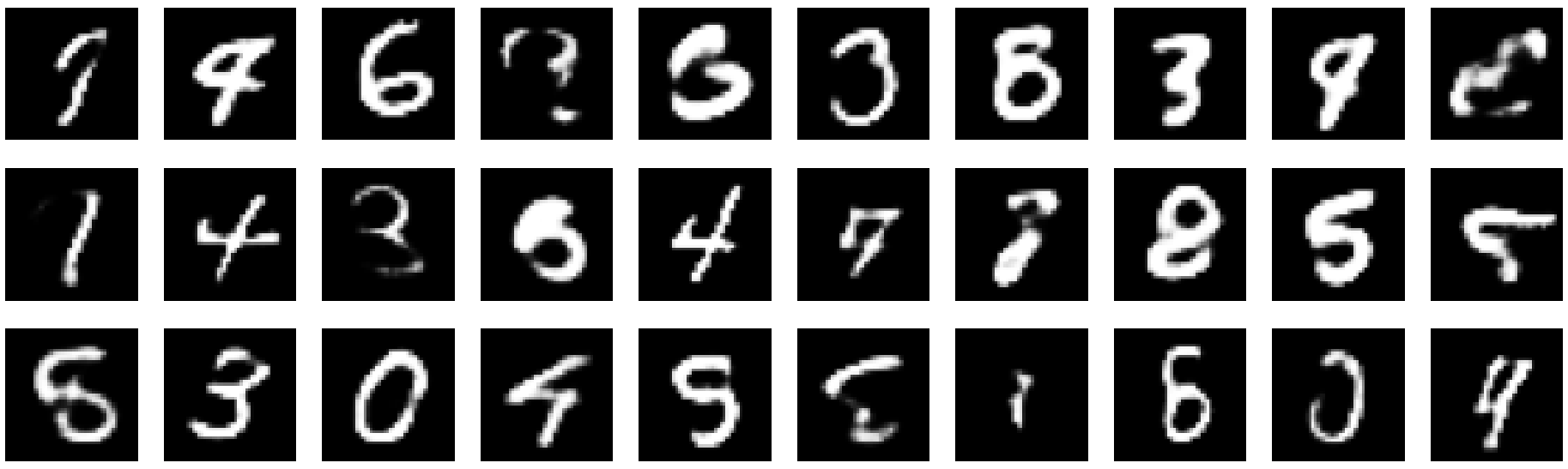

**Figure 6.** Parameters of 30 Products of Continuous Bernoullis obtained from 30 latent draws.

```
        xs = self.shared(xs)
        return Ind(Normal(self.mus(xs), self.sigmas(xs)), 1)
```

Recall that optimizing with respect to $\boldsymbol{\phi}$ has the interpretation of minimizing the (reverse) KL between $q_{\boldsymbol{\phi}}(\boldsymbol{z}_i|\,\boldsymbol{x}_i)$ and $p_{\boldsymbol{\theta}}(\boldsymbol{z}_i|\,\boldsymbol{x}_i)$. Since $p_{\boldsymbol{\theta}}(\boldsymbol{z}_i|\,\boldsymbol{x}_i)$ is likely to be multi-modal due to nonlinearlity in the generative model, minimizing the reverse KL leads to *zero forcing* behavior of $q_{\boldsymbol{\phi}}(\boldsymbol{z}_i|\,\boldsymbol{x}_i)$, i.e., it locks onto a single mode of $p_{\boldsymbol{\theta}}(\boldsymbol{z}_i|\,\boldsymbol{x}_i)$. One could technically use a multi-modal approximate posterior (e.g., see [4] for reparametrization trick for mixture densities), leading to a tighter ELBO. However, it turns out that using Gaussians with diagonal covariance can already lead to satisfactory performance (in terms of quality of generated outputs) in practice.

### 5.3.3 Results

We trained[18] a VAE on the normalized MNIST dataset by optimizing a mini-batch estimate of ELBO via gradient ascent until convergence. For training hyperparameters, we used a latent dimension of 20, a mini-batch size of 100, a dropout rate of $p = 0.1$ and Adam with a learning rate of 3e-4.

Figure 5 shows the estimated ELBO on the test set across training. Training was halted after estimated ELBO plateaus. We plotted the parameters of 30 Product of Continuous Bernoullis obtained from 30 latent draws from $p(\boldsymbol{z})$ in Figure 6, and plotted one sample from each Product of Continuous Bernoullis in Figure 7.

Compared to the original MNIST digits, sampled digits have a lot more noise littered across the image, which can be attributed to the model's inability to capture correlation between pixels. The quality of generation is also highly variable. For example, in Figure 6, the digit on the first row and second column has excellent quality while the digit on the first row and fourth column doesn't look like a digit. Overall, the generative model seems to have picked up the distribution of digits, but there's clearly room for improvement.

18. Because AEVB for VAE is identical to AEVB for FA, we simply use the estimator derived in Section 5.2.3.

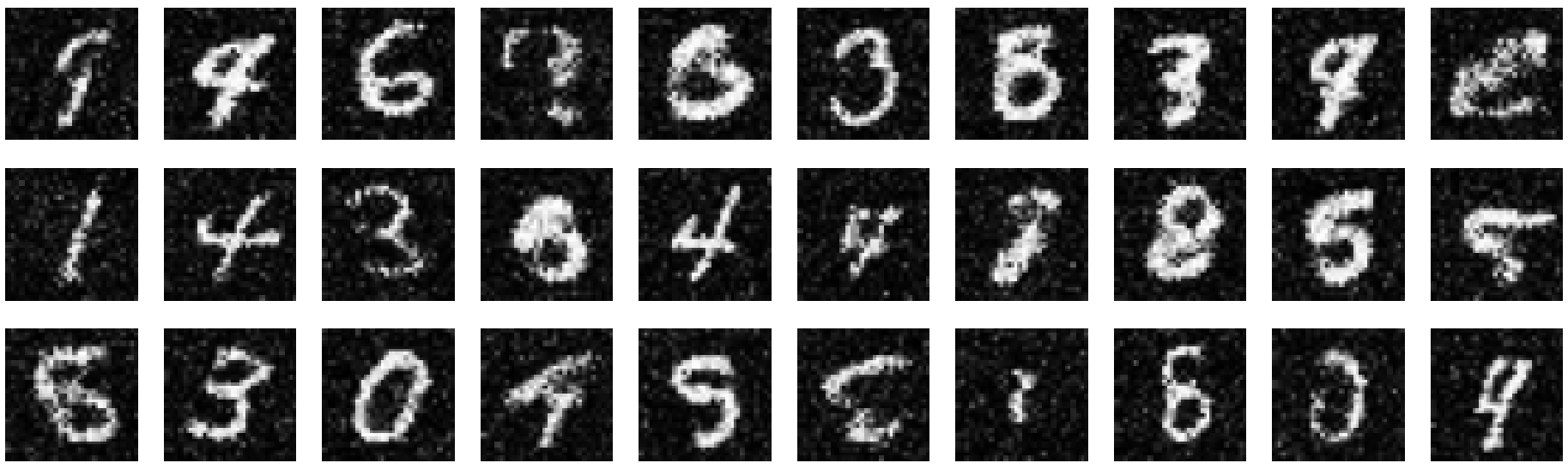

**Figure 7.** Samples from 30 Product of Continuous Bernoullis whose parameters are plotted in Figure 6.

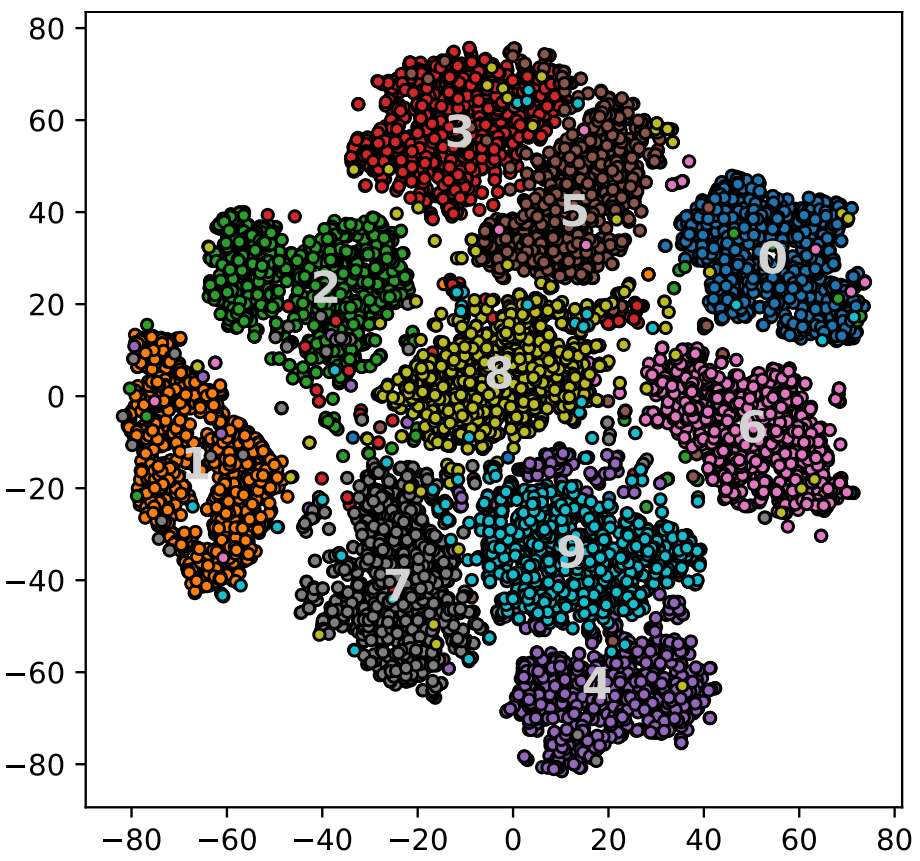

**Figure 8.** 2-dimensional t-SNE projection of latent means (20 dimensional) obtained from test set. Digit labels (grey) are placed at the median of latent means of each class. We see that the resulting clusters correspond nicely to clusters we had in mind as humans.

We also visualized the structure of the latent space in Figure 8. We computed means of $q_\phi(\boldsymbol{z}_i|\, \boldsymbol{x}_i)$ for all $\boldsymbol{x}_i$ in the test set, applied scikit-learn's implementation of t-SNE ([20]) to project these latent means to 2 dimensions, and finally scattered and colored the output vectors. We see that the resulting clusters correspond nicely to clusters we had in mind as humans.

## 5.4 Conditional VAE

### 5.4.1 Generative model

While VAE can generate novel data with good quality, we might want to additionally condition our model to generate novel data based on some other observed quantity $\boldsymbol{y}_i$. For a concrete example, in the context of normalized MNIST, we might to generate new digits from a specific class – in this case $\boldsymbol{y}_i$ would represent the one-hot class label. More formally, this is the problem of modelling the conditional density $p_{\boldsymbol{\theta}}(\boldsymbol{x}_i|\, \boldsymbol{y}_i)$ using a latent variable model. Such a model has a joint density of $p_{\boldsymbol{\theta}}(\boldsymbol{x}_i, \boldsymbol{z}_i|\, \boldsymbol{y}_i)$, and the full decomposition of the joint is $p_{\boldsymbol{\theta}}(\boldsymbol{x}_i|\, \boldsymbol{y}_i, \boldsymbol{z}_i) p_{\boldsymbol{\theta}}(\boldsymbol{z}_i|\, \boldsymbol{y}_i)$, which is exactly the model specified in the Conditional VAE (CVAE) [19]:

$$\begin{aligned} \boldsymbol{x}_i &\sim \text{Product-Of-Continuous-Bernoullis}(\boldsymbol{p}_{\boldsymbol{\theta}}(\boldsymbol{y}_i, \boldsymbol{z}_i)) \\ \boldsymbol{z}_i &\sim \mathcal{N}(\boldsymbol{\mu}_{\boldsymbol{\theta}}(\boldsymbol{y}_i), \boldsymbol{\sigma}_{\boldsymbol{\theta}}(\boldsymbol{y}_i)). \end{aligned}$$

It is worth noting that this is not the only reasonable model. Here are two other models that assume some independence instead of using the full decomposition. One option is to only condition $\boldsymbol{z}_i$ on $\boldsymbol{y}_i$:

$$\begin{aligned} \boldsymbol{x}_i &\sim \text{Product-Of-Continuous-Bernoullis}(\boldsymbol{p}_{\boldsymbol{\theta}}(\boldsymbol{z}_i)) \\ \boldsymbol{z}_i &\sim \mathcal{N}(\boldsymbol{\mu}_{\boldsymbol{\theta}}(\boldsymbol{y}_i), \boldsymbol{\sigma}_{\boldsymbol{\theta}}(\boldsymbol{y}_i)). \end{aligned}$$

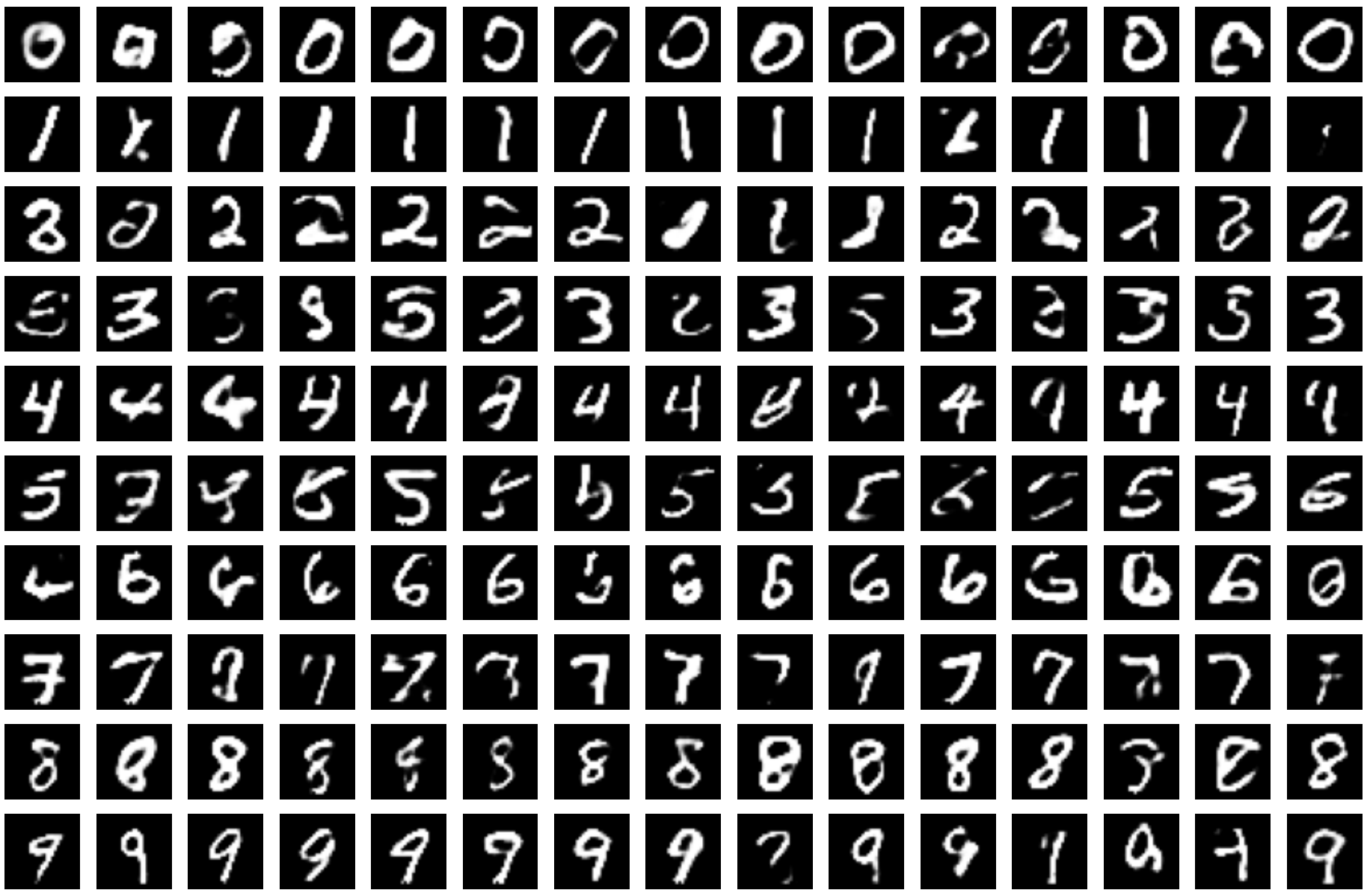

**Figure 9.** Generated images by Conditional VAE. Each row contains all generations from the same label.

The other option is to only condition $\boldsymbol{x}_i$ on $\boldsymbol{y}_i$:

$$\begin{aligned} \boldsymbol{x}_i &\sim \text{Product-Of-Bernoullis}(\boldsymbol{p}_{\boldsymbol{\theta}}(\boldsymbol{y}_i, \boldsymbol{z}_i)) \\ \boldsymbol{z}_i &\sim \mathcal{N}(0, \boldsymbol{I}_L). \end{aligned}$$

In this section, we will focus on the first model that does not assume independence, though it's straightforward to extend the training procedure to the other two models.

### 5.4.2 Approximate posterior

Similar to how we chose $q_{\boldsymbol{\phi}}(\boldsymbol{z}_i|\,\boldsymbol{x}_i)$ to be a diagonal Gaussian for VAE, here for CVAE we choose $q_{\boldsymbol{\phi}}(\boldsymbol{z}_i|\,\boldsymbol{x}_i, \boldsymbol{y}_i)$ to be a diagonal Gaussian.

### 5.4.3 Estimator for per-example ELBO

Following the approach in Section 5.2.3, we first derive an unbiased, low-variance estimator of the per-example ELBO whose randomness does not depend on the parameters. The per-example ELBO for CVAE is almost identical to that of FA and VAE, except that the generative and inference components must now be conditioned by $\boldsymbol{y}_i$ as follows:

$$\mathbb{E}_{\boldsymbol{z}_i \sim q_{\boldsymbol{\phi}}(\boldsymbol{z}_i|\boldsymbol{x}_i, \boldsymbol{y}_i)}[\log p_{\boldsymbol{\theta}}(\boldsymbol{x}_i, \boldsymbol{z}_i|\,\boldsymbol{y}_i) - \log q_{\boldsymbol{\phi}}(\boldsymbol{z}_i|\,\boldsymbol{x}_i, \boldsymbol{y}_i)].$$

Therefore, the estimator is simply:

$$\log p_{\boldsymbol{\theta}}(\boldsymbol{x}_i|\,\boldsymbol{z}_i^s, \boldsymbol{y}_i) - D_{\mathbb{KL}}(q_{\boldsymbol{\phi}}(\boldsymbol{z}_i|\,\boldsymbol{x}_i, \boldsymbol{y}_i) \,\|\, p_{\boldsymbol{\theta}}(\boldsymbol{z}_i|\,\boldsymbol{y}_i)) \quad (\boldsymbol{z}_i^s = \boldsymbol{\mu}_{\boldsymbol{\phi}}(\boldsymbol{x}_i, \boldsymbol{y}_i) + \boldsymbol{\sigma}_{\boldsymbol{\phi}}(\boldsymbol{x}_i, \boldsymbol{y}_i)\boldsymbol{\varepsilon}_i, \boldsymbol{\varepsilon}_i \sim \mathcal{N}(0, \boldsymbol{I}_L)).$$

### 5.4.4 Results

We trained a CVAE on the normalized MNIST dataset by optimizing a mini-batch estimate of ELBO via gradient ascent until convergence. We used the same hyperparameters for training VAE.

After training, we conditionally generated new digits from each class by ancestral sampling of $p_{\boldsymbol{\phi}}(\boldsymbol{z}_i|\,\boldsymbol{y}_i)$ and then $p_{\boldsymbol{\phi}}(\boldsymbol{x}_i, \boldsymbol{z}_i|\,\boldsymbol{y}_i)$. The generations are shown in Figure 9, where each row contains all generations from the same label. The fact that this works well shows that the digit label was indeed exploited by the model for reconstruction during training; if we instead passed in random values for $\boldsymbol{y}_i$ instead of ground-truth labels, the model would have learned that the digit label is not helpful for reconstruction and we would not be able to achieve such results.

## 5.5 Gaussian Mixture VAE

### 5.5.1 Generative model

The generative part of the Gaussian Mixture VAE (GMVAE) [18] for binary MNIST is

$$\begin{aligned}\boldsymbol{x}_i &\sim \text{Product-Of-Bernoullis}(\boldsymbol{p_\theta}(\boldsymbol{z}_i))\\ \boldsymbol{z}_i &\sim \mathcal{N}(\boldsymbol{\mu_\theta}(\boldsymbol{y}_i), \boldsymbol{\sigma_\theta}(\boldsymbol{y}_i))\\ \boldsymbol{y}_i &\sim \text{One-Hot-Categorical}(\boldsymbol{\pi})\end{aligned}$$

where both $\boldsymbol{y}_i$, a $C$-dimensional one-hot vector, and $\boldsymbol{z}_i \in \mathbb{R}^L$ are latent variables, and $\boldsymbol{x}_i \in \{0,1\}^D$ is the observed variable. It may seem strange that we now have two latent variables instead of one, but we could have treated them as a single random variable $\boldsymbol{y}_i \,\|\, \boldsymbol{z}_i$ of dimension $L + D$, where $\|$ denotes concatenation. Additionally, $\boldsymbol{\pi}$, chosen to be a length $C$ vector $(1/C, \ldots, 1/C)$, lives in the $C$-dimensional simplex; $\boldsymbol{\mu_\theta}$ and $\boldsymbol{\sigma_\theta}$ are both linear transformations (so multiplying with a one-hot vector $\boldsymbol{y}_i$ effectively selects a column from their corresponding matrices) except that $\boldsymbol{\sigma_\theta}$ is handled with care such that its output is non-negative; $p_{\boldsymbol{\theta}}$ is a multi-layer neural network that maps the latent variable $\boldsymbol{z}_i$ into success probabilities of the Product-Of-Bernoullis distribution.

While this generative model is interpretable, it is not the only such model. One can argue that the following model, which resembles the third model in Section 5.4, is also interpretable:

$$\begin{aligned}\boldsymbol{x}_i &\sim \text{Product-Of-Bernoullis}(\boldsymbol{p_\theta}(\boldsymbol{y}_i, \boldsymbol{z}_i))\\ \boldsymbol{z}_i &\sim \mathcal{N}(0, \boldsymbol{I}_L)\\ \boldsymbol{y}_i &\sim \text{One-Hot-Categorical}(\boldsymbol{\pi}),\end{aligned}$$

where $\boldsymbol{y}_i$ and $\boldsymbol{z}_i$ can be taken intuitively as class label and stylistic features of handwritten digits respectively. In fact, this is the M2 model for semi-supervised classification discussed in [10]. For a short empirical discussion as to why this model doesn't work well for unsupervised clustering and potential fixes, see [18].

### 5.5.2 Approximate posterior

The true joint posterior $p_{\boldsymbol{\theta}}(\boldsymbol{y}_i, \boldsymbol{z}_i | \boldsymbol{x}_i)$ factors into $p_{\boldsymbol{\theta}}(\boldsymbol{y}_i | \boldsymbol{x}_i) p_{\boldsymbol{\theta}}(\boldsymbol{z}_i | \boldsymbol{x}_i, \boldsymbol{y}_i)$ and we can decide on a family of distributions for each of them separately. The decision for $p_{\boldsymbol{\theta}}(\boldsymbol{y}_i | \boldsymbol{x}_i)$ is straightforward: it is a one-hot categorical distribution whose parameters are related to $\boldsymbol{x}_i$ in a highly nonlinear fashion. We therefore model this exactly using

$$q_{\boldsymbol{\phi}}(\boldsymbol{y}_i | \boldsymbol{x}_i) = \text{OneHotCategorical}(\boldsymbol{\pi_\phi}(\boldsymbol{x}_i))$$

where $\boldsymbol{\pi_\phi}$ is a flexible neural network that maps from $\mathbb{R}^D$ to the $C$-dimensional simplex. For $p_{\boldsymbol{\theta}}(\boldsymbol{z}_i | \boldsymbol{x}_i, \boldsymbol{y}_i)$, we choose $q_{\boldsymbol{\phi}}(\boldsymbol{z}_i | \boldsymbol{x}_i, \boldsymbol{y}_i)$ to be the diagonal Gaussian family as for VAE. While the alternative factorization $q_{\boldsymbol{\phi}}(\boldsymbol{y}_i, \boldsymbol{z}_i | \boldsymbol{x}_i) = q_{\boldsymbol{\phi}}(\boldsymbol{y}_i | \boldsymbol{x}_i, \boldsymbol{z}_i) q_{\boldsymbol{\phi}}(\boldsymbol{z}_i | \boldsymbol{x}_i)$ is also valid, it doesn't give us the ability to cluster observations $\boldsymbol{x}_i$, since evaluating $q_{\boldsymbol{\phi}}(\boldsymbol{y}_i | \boldsymbol{x}_i)$ would require evaluating the integral

$$q_{\boldsymbol{\phi}}(\boldsymbol{y}_i | \boldsymbol{x}_i) = \int q_{\boldsymbol{\phi}}(\boldsymbol{y}_i | \boldsymbol{x}_i, \boldsymbol{z}_i) q_{\boldsymbol{\phi}}(\boldsymbol{z}_i | \boldsymbol{x}_i)\, d\boldsymbol{z}_i.$$

### 5.5.3 Estimator for per-example ELBO

**Estimator 1: Marginalization of $\boldsymbol{y}_i$**

Following the approach in Section 5.2.3, we first derive an unbiased, low-variance estimator of the per-example ELBO whose randomness does not depend on the parameters. We marginalize out $\boldsymbol{y}_i$ by taking the advantage of the fact that it's discrete, and only reparametrize $\boldsymbol{z}_i$:

$$
\begin{aligned}
&\mathbb{E}_{q_{\boldsymbol{\phi}}(\boldsymbol{y}_i,\boldsymbol{z}_i|\boldsymbol{x}_i)}\Big[\log p_{\boldsymbol{\theta}}(\boldsymbol{x}_i,\boldsymbol{y}_i,\boldsymbol{z}_i) - \log q_{\boldsymbol{\phi}}(\boldsymbol{y}_i,\boldsymbol{z}_i|\,\boldsymbol{x}_i)\Big] \quad \text{(per-example ELBO)} \\
=\ &\mathbb{E}_{q_{\boldsymbol{\phi}}(\boldsymbol{y}_i|\boldsymbol{x}_i)q_{\boldsymbol{\phi}}(\boldsymbol{z}_i|\boldsymbol{y}_i,\boldsymbol{x}_i)}\Big[\log p_{\boldsymbol{\theta}}(\boldsymbol{x}_i|\,\boldsymbol{z}_i) + \log p_{\boldsymbol{\theta}}(\boldsymbol{z}_i|\,\boldsymbol{y}_i) + \log p(\boldsymbol{y}_i) - \log q_{\boldsymbol{\phi}}(\boldsymbol{y}_i|\,\boldsymbol{x}_i) - \log q_{\boldsymbol{\phi}}(\boldsymbol{z}_i|\,\boldsymbol{y}_i,\boldsymbol{x}_i)\Big] \\
=\ &\mathbb{E}_{q_{\boldsymbol{\phi}}(\boldsymbol{y}_i|\boldsymbol{x}_i)}\Big[\mathbb{E}_{q_{\boldsymbol{\phi}}(\boldsymbol{z}_i|\boldsymbol{y}_i,\boldsymbol{x}_i)}[\log p_{\boldsymbol{\theta}}(\boldsymbol{x}_i|\,\boldsymbol{z}_i)] + \mathbb{E}_{q_{\boldsymbol{\phi}}(\boldsymbol{z}_i|\boldsymbol{y}_i,\boldsymbol{x}_i)}[\log p_{\boldsymbol{\theta}}(\boldsymbol{z}_i|\,\boldsymbol{y}_i) - \log q_{\boldsymbol{\phi}}(\boldsymbol{z}_i|\,\boldsymbol{y}_i,\boldsymbol{x}_i)]\Big] \\
&+\mathbb{E}_{q_{\boldsymbol{\phi}}(\boldsymbol{y}_i|\boldsymbol{x}_i)}\Big[\log p(\boldsymbol{y}_i) - \log q_{\boldsymbol{\phi}}(\boldsymbol{y}_i|\,\boldsymbol{x}_i)\Big] \\
\backsimeq\ &\mathbb{E}_{q_{\boldsymbol{\phi}}(\boldsymbol{y}_i|\boldsymbol{x}_i)}\Big[\underbrace{\log p_{\boldsymbol{\theta}}(\boldsymbol{x}_i|\,\boldsymbol{z}_i^s)}_{\boldsymbol{z}_i^s \text{ is a reparametrized sample from } q_{\boldsymbol{\phi}}(\boldsymbol{z}_i|\boldsymbol{y}_i,\boldsymbol{x}_i)} - D_{\mathbb{KL}}(q_{\boldsymbol{\phi}}(\boldsymbol{z}_i|\,\boldsymbol{y}_i,\boldsymbol{x}_i)|\,p_{\boldsymbol{\theta}}(\boldsymbol{z}_i|\,\boldsymbol{y}_i))\Big] - D_{\mathbb{KL}}(q_{\boldsymbol{\phi}}(\boldsymbol{y}_i|\boldsymbol{x}_i)|\,p(\boldsymbol{y}_i)) \\
=\ &\left(\sum_{\boldsymbol{y}_i} q_{\boldsymbol{\phi}}(\boldsymbol{y}_i|\,\boldsymbol{x}_i)\cdot[p_{\boldsymbol{\theta}}(\boldsymbol{x}_i|\,\boldsymbol{z}_i^s) - D_{\mathbb{KL}}(q_{\boldsymbol{\phi}}(\boldsymbol{z}_i|\,\boldsymbol{y}_i,\boldsymbol{x}_i)|\,p_{\boldsymbol{\theta}}(\boldsymbol{z}_i|\,\boldsymbol{y}_i))]\right) - D_{\mathbb{KL}}(q_{\boldsymbol{\phi}}(\boldsymbol{y}_i|\,\boldsymbol{x}_i)|\,p(\boldsymbol{y}_i)) \qquad (5)
\end{aligned}
$$

where, in the second step, we distributed the expectation operator with respect to $q(\boldsymbol{z}_i|\,\boldsymbol{y}_i,\boldsymbol{x}_i)$ to different terms and, in the third step, we sampled $\boldsymbol{z}_i^s$ from $q_{\boldsymbol{\phi}}(\boldsymbol{z}_i|\,\boldsymbol{y}_i,\boldsymbol{x}_i)$ using the reparametrization trick to ensure that the randomness of the estimator does not depend on $\boldsymbol{\phi}$. Since $p(\boldsymbol{y}_i)$ has no trainable parameters, $-D_{\mathbb{KL}}(q(\boldsymbol{y}_i|\boldsymbol{x}_i)|\,p(\boldsymbol{y}_i))$ can be written instead as $\mathbb{H}(q(\boldsymbol{y}_i|\,\boldsymbol{x}_i))$, the conditional entropy of the unsupervised classifer, plus a constant which can be ignored. In PyTorch, Equation 5.5.3.1 can be conveniently vectorized and evaluated as follows:

```
class AEVB(nn.Module):
    # ...
    def step(self, xs):
        # xs shape: (batch size, 28 * 28)

        q_y = self.q_y_given_x(xs)
        cond_ent = q_y.entropy()

        # ========== code below vectorizes marginalization =========

        bs = xs.size(0)
        ys = torch.eye(self.y_dim)[torch.tensor(range(self.y_dim)).repeat_interleave(bs)]
        xs = xs.repeat(self.y_dim, 1)

        post_over_zs = self.q_z(xs, ys)
        zs = post_over_zs.rsample()

        other = self.p_x(zs).log_prob(xs) - kl_divergence(post_over_zs, self.p_z(ys))

        other = other.reshape(self.y_dim, bs).T
        other = (q_y.probs * other).sum(dim=1)

        # ========== code above vectorizes marginalization =========

        per_example_elbos = other + cond_ent  # values of estimators of per-example ELBOs
        mini_batch_elbo = per_example_elbos.mean()
        loss = - mini_batch_elbo

        self.opt.zero_grad()
        loss.backward()
        self.opt.step()
    # ...
```

### Estimator 2: Gumbel-Softmax trick

It has been shown [5, 14] that one can obtain a reparametrized sample of $\boldsymbol{y}_i$ by

$$\boldsymbol{y}_i = \text{onehot}(\text{argmax}(\text{logits} + \boldsymbol{g}_i))$$

where $\boldsymbol{g}_i = (g_{i,1}, \ldots, g_{i,c})$ are i.i.d. samples from Gumbel$(0, 1)$ distribution and logits is a length-$C$ vector containing normalized or unnormalized log probabilities of the original one-hot categorical distribution $q_{\boldsymbol{\phi}}(\boldsymbol{y}_i | \boldsymbol{x}_i)$. However, the resulting sample cannot be differentiated through due to the argmax operation, since argmax has a gradient of zero at non-zero inputs. Instead,, [7, 13] replaces argmax with softmax to make it differentiable:

$$\tilde{\boldsymbol{y}}_i = \text{softmax}((\text{logits} + \boldsymbol{g}_i) / \tau)$$

where $\tau > 0$ is called the temperature: when it's low, the distribution of $\tilde{\boldsymbol{y}}_i$ approaches the distribution of $\boldsymbol{y}_i$; when it's high, $\tilde{\boldsymbol{y}}_i$ would lie closer to the center of the simplex. The resulting distribution is called the Gumbel-Softmax distribution [7] or the Concrete distribution [13], and the procedure of using a differentiable but approximate sample of $\boldsymbol{y}_i$ is called the Gumbel-Softmax trick or estimator. For a straight-through variant, see the original paper [7].

The obvious consequence of using softmax is that the samples are no longer one-hot, which may seem counter-intuitive. The key to understanding this is by recognizing that the one-hot categorical distribution and the Gumbel-Softmax distribution are very similar (in terms of expectation and how samples are close to the corners of the simplex) at low temperature (e.g., at $\tau \leqslant 1$) under the same parameter values so that they can be used as drop-in substitutions for one another, depending on the context. One can therefore see these them as separate "heads" attached to the same vector of logits. Below we discuss two such contexts when estimating the ELBO.

**One context.** We can derive a *slightly biased* (due to replacing argmax with softmax) estimator[19] of per-example the ELBO by first drawing reparametrized samples $(\tilde{\boldsymbol{y}}_i^s, \boldsymbol{z}_i^s)$ of the expectation via ancestral sampling, and then evaluating the interior of the expectation:

$$\begin{aligned} & \mathbb{E}_{q_{\boldsymbol{\phi}}(\boldsymbol{y}_i|\boldsymbol{x}_i) q_{\boldsymbol{\phi}}(\boldsymbol{z}_i|\boldsymbol{y}_i,\boldsymbol{x}_i)}[\log p_{\boldsymbol{\theta}}(\boldsymbol{x}_i|\boldsymbol{z}_i) + \log p_{\boldsymbol{\theta}}(\boldsymbol{z}_i|\boldsymbol{y}_i) + \log p_{\boldsymbol{\theta}}(\boldsymbol{y}_i) - \log q_{\boldsymbol{\phi}}(\boldsymbol{y}_i|\boldsymbol{x}_i) - \log q_{\boldsymbol{\phi}}(\boldsymbol{z}_i|\boldsymbol{y}_i,\boldsymbol{x}_i)] \\ \backsimeq\ & \underbrace{\log p_{\boldsymbol{\theta}}(\boldsymbol{x}_i|\tilde{\boldsymbol{y}}_i^s, \boldsymbol{z}_i^s) + \log p_{\boldsymbol{\theta}}(\boldsymbol{z}_i^s|\tilde{\boldsymbol{y}}_i^s) + \log p_{\boldsymbol{\theta}}(\tilde{\boldsymbol{y}}_i^s) - \log q_{\boldsymbol{\phi}}(\tilde{\boldsymbol{y}}_i^s|\boldsymbol{x}_i^s) - \log q_{\boldsymbol{\phi}}(\boldsymbol{z}_i^s|\tilde{\boldsymbol{y}}_i^s, \boldsymbol{x}_i)}_{(\tilde{\boldsymbol{y}}_i^s, \boldsymbol{z}_i^s) \text{ is a reparametrized sample from } q(\tilde{\boldsymbol{y}}_i|\boldsymbol{x}_i) q(\boldsymbol{z}_i|\tilde{\boldsymbol{y}}_i, \boldsymbol{x}_i)} \end{aligned}$$

where in particular $\tilde{\boldsymbol{y}}_i^s$ is sampled using the Gumbel-Softmax trick. The immediate problem is that $p(\boldsymbol{y}_i)$ and $q(\boldsymbol{y}_i|\boldsymbol{x}_i)$ are by default categorical distributions and hence cannot evaluate the log probability of sampled $\tilde{\boldsymbol{y}}_i^s$ since $\tilde{\boldsymbol{y}}_i^s$ is *not* one-hot. The solution is to use the Gumbel-Softmax head for $p(\boldsymbol{y}_i)$ and $q(\boldsymbol{y}_i|\boldsymbol{x}_i)$. In PyTorch, these two heads can be implemented as:

```
class q_y_class(nn.Module):

    def __init__(self, x_dim, y_dim):
        super().__init__()
        self.x_dim = x_dim
        self.y_dim = y_dim
        self.logits = nn.Sequential(
            # linear layers and activations
        )

    def forward(self, xs, head):
        # xs shape: (batch size, 28 * 28)
        if head == "one_hot_categorical":
            return OneHotCat(logits=self.logits(xs))
        elif head == "gumbel_softmax":
            return RelaxedOneHotCat(logits=self.logits(xs), temperature=torch.tensor([0.5]))
```

After training, when generating $\boldsymbol{x}_i$ from samples of $p(\boldsymbol{y}_i)$ and doing classification using $q(\boldsymbol{y}_i|\boldsymbol{x}_i)$, we can switch back to using the One-Hot-Categorical head, since they are more interpretable for these tasks and represent the what we truly want, i.e., discrete latent variable. Note that distributions $p(\boldsymbol{z}_i|\boldsymbol{y}_i)$ and $q(\boldsymbol{z}_i|\boldsymbol{y}_i, \boldsymbol{x}_i)$ do not suffer from this problem because, as neural networks, they can process $\tilde{\boldsymbol{y}}_i^s$ as long as it consists of real entries. And since $\tilde{\boldsymbol{y}}_i^s$ can be very close to one-hot (though never strictly) during training, these neural networks would generalize to one-hot samples for purposes after training, e.g., using $p(\boldsymbol{z}_i|\boldsymbol{y}_i)$ for conditional generation of $\boldsymbol{x}_i$.

19. Obviously, this estimator has higher variance than the estimator that uses marginalization, but evaluating this estimator would have better time complexity than doing marginalization, especially when $C$ is large.

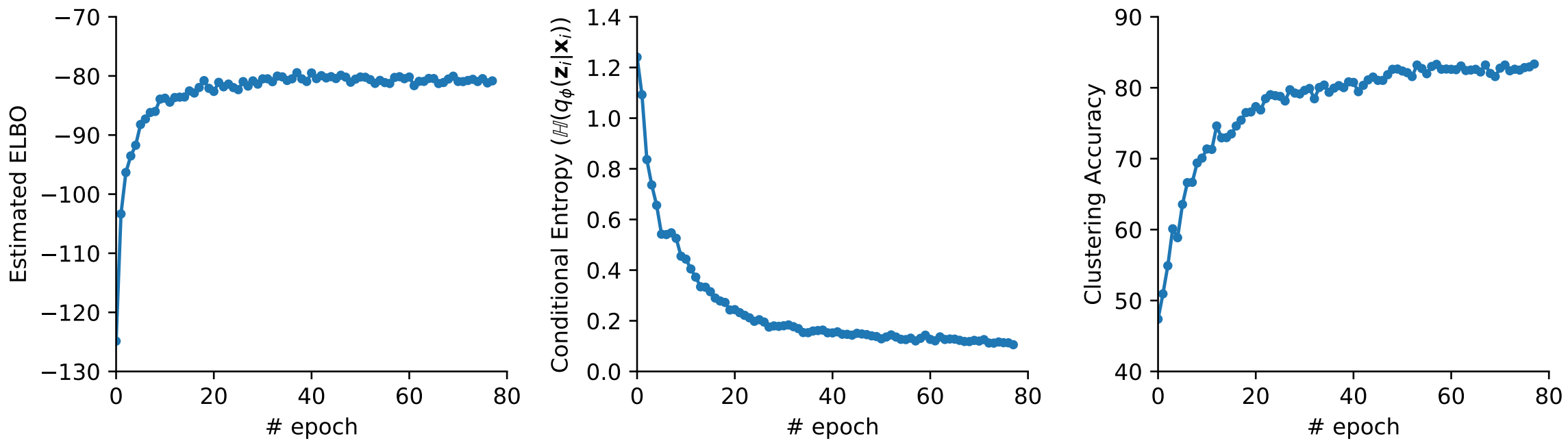

**Figure 10.** Test set performance of a Gaussian Mixture VAE with $C=10$ trained on binary MNIST. Estimated ELBO increases over time as expected. Conditional entropy and clustering accuracy indicates that the model learned to exploit the categorical latent variable for encoding information efficiently.

**Another context.** There are many ways to form an estimator of the per-example ELBO. Apart from the one just discussed, a popular implementation[20] decomposes the per-example ELBO as

$$\begin{aligned} & \mathbb{E}_{q_{\boldsymbol{\phi}}(\boldsymbol{y}_i|\boldsymbol{x}_i)q_{\boldsymbol{\phi}}(\boldsymbol{z}_i|\boldsymbol{y}_i,\boldsymbol{x}_i)}\Big[\log p_{\boldsymbol{\theta}}(\boldsymbol{x}_i|\boldsymbol{z}_i)+\log p_{\boldsymbol{\theta}}(\boldsymbol{z}_i|\boldsymbol{y}_i)+\log p_{\boldsymbol{\theta}}(\boldsymbol{y}_i)-\log q_{\boldsymbol{\phi}}(\boldsymbol{y}_i|\boldsymbol{x}_i)-\log q_{\boldsymbol{\phi}}(\boldsymbol{z}_i|\boldsymbol{y}_i,\boldsymbol{x}_i)\Big] \\ = \; & \mathbb{E}_{q_{\boldsymbol{\phi}}(\boldsymbol{y}_i|\boldsymbol{x}_i)}\Big[\mathbb{E}_{q_{\boldsymbol{\phi}}(\boldsymbol{z}_i|\boldsymbol{y}_i,\boldsymbol{x}_i)}[\log p_{\boldsymbol{\theta}}(\boldsymbol{x}_i|\boldsymbol{z}_i)]+\mathbb{E}_{q_{\boldsymbol{\phi}}(\boldsymbol{z}_i|\boldsymbol{y}_i,\boldsymbol{x}_i)}[\log p_{\boldsymbol{\theta}}(\boldsymbol{z}_i|\boldsymbol{y}_i)-\log q_{\boldsymbol{\phi}}(\boldsymbol{z}_i|\boldsymbol{y}_i,\boldsymbol{x}_i)]\Big] \\ & +\mathbb{E}_{q_{\boldsymbol{\phi}}(\boldsymbol{y}_i|\boldsymbol{x}_i)}\Big[\log p(\boldsymbol{y}_i)-\log q_{\boldsymbol{\phi}}(\boldsymbol{y}_i|\boldsymbol{x}_i)\Big] \\ \backsimeq \; & \log p_{\boldsymbol{\theta}}(\boldsymbol{x}_i|\boldsymbol{z}_i^s)-D_{\mathbb{KL}}(q_{\boldsymbol{\phi}}(\boldsymbol{z}_i|\tilde{\boldsymbol{y}}_i^s,\boldsymbol{x}_i)|\,p_{\boldsymbol{\theta}}(\boldsymbol{z}_i|\tilde{\boldsymbol{y}}_i^s))-D_{\mathbb{KL}}(q_{\boldsymbol{\phi}}(\boldsymbol{y}_i|\boldsymbol{x}_i)|\,p(\boldsymbol{y}_i)), \end{aligned}$$

where $(\tilde{\boldsymbol{y}}_i^s,\boldsymbol{z}_i^s)$ is again a reparametrized sample from $q_{\boldsymbol{\phi}}(\tilde{\boldsymbol{y}}_i|\boldsymbol{x}_i)q_{\boldsymbol{\phi}}(\boldsymbol{z}_i|\tilde{\boldsymbol{y}}_i,\boldsymbol{x}_i)$. In this context, since the KL between the two categorical distributions is computed analytically, there's no need to use the Gumbel-Softmax head for computing log probabilities of $\tilde{\boldsymbol{y}}_i^s$.

### 5.5.4 Results

We trained a GMVAE on binary MNIST by optimizing a mini-batch estimate of ELBO via gradient ascent until convergence. In particular, to reduce variance as much as possible, we use the marginalizing estimator as described in Section 5.5.3. We used $C=10$ since there are 10 classes digits in MNIST. For training hyperparameters, we used the Glorot-Normal [3] initializer for weights and zeros for biases[21], a mini-batch size of 100 and Adam with a learning rate of 0.001.

Figure 10 shows the estimated ELBO, conditional entropy and clustering accuracy on the entire test set after each epoch of training. Unsurprisingly, ELBO improves over time and plateaus. More interestingly, we see that the conditional entropy decreases and clustering accuracy increases over time. Since the model imposes a uniform prior over the classes, the model would only benefit from decreasing conditional entropy if it finds encoding information in $q_{\boldsymbol{\phi}}(\boldsymbol{y}_i|\boldsymbol{x}_i)$ helpful for reconstruction. At least for MNIST (for another such dataset, see Figure 4 of [21]), it turns out that the model clustered digits like a human would, but for datasets without obvious subclasses the resulting clusters might be a lot less semantically meaningful and predictable.

Apart from the training statistics, we can also conditionally generate new digits (Figure 11) from the trained model obtained through ancestral sampling of $p_{\boldsymbol{\theta}}(\boldsymbol{z}|\boldsymbol{y})$ and then $p_{\boldsymbol{\theta}}(\boldsymbol{x}|\boldsymbol{z})$. Different clusters learned to generated distinct digits, though a few clusters do overlap. This observation couples with a good clustering accuracy: if all digits from the a specific class are encoded to a specific cluster during training, then the model would also know how to decode them back to digits from that same class with the help of the reconstruction loss.

20. Code: `https://github.com/jariasf/GMVAE`

21. Following the recommentation here: `https://github.com/RuiShu/vae-clustering/issues/10`

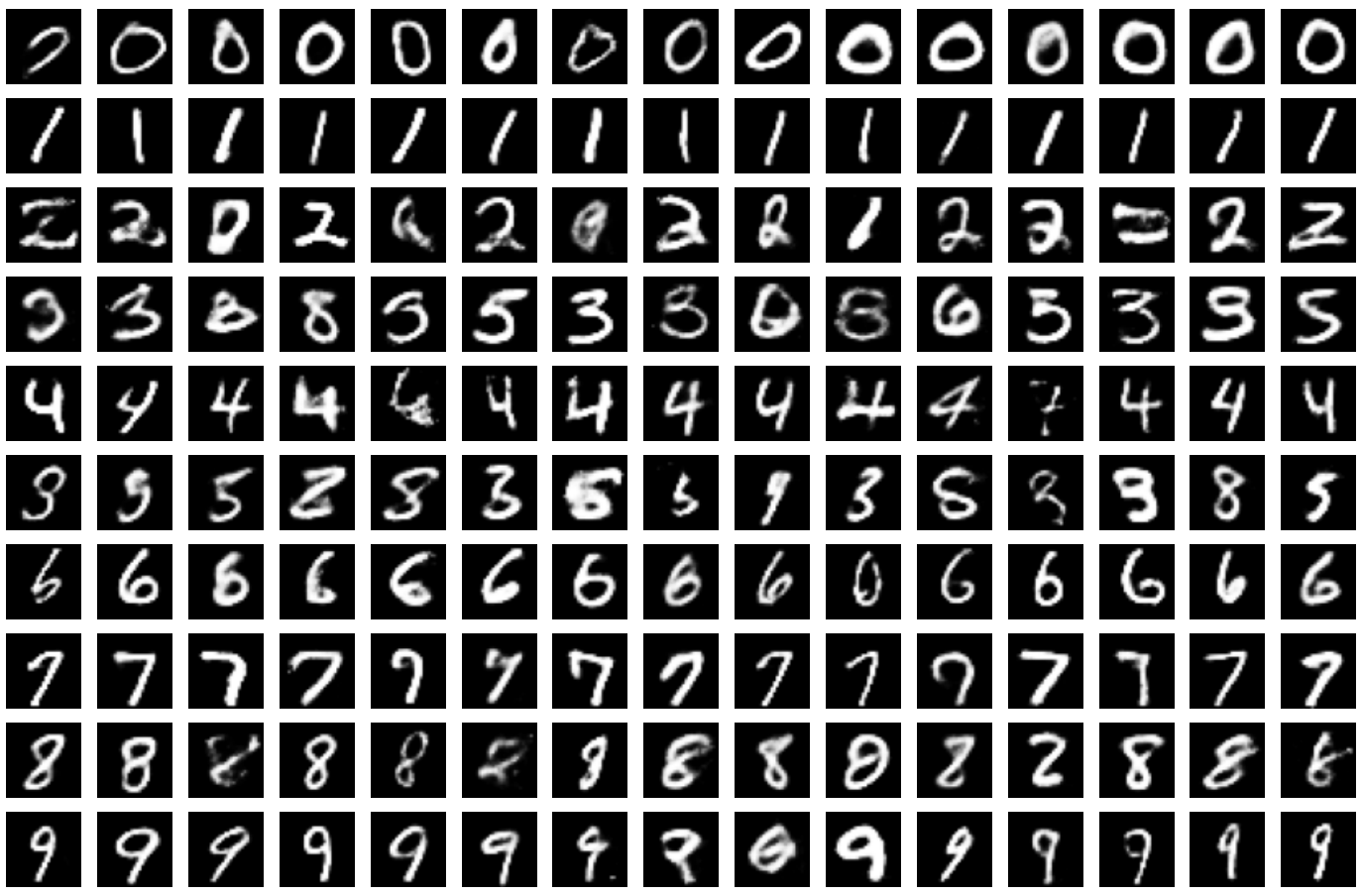

**Figure 11.** Conditional generations of a Gaussian Mixture VAE with $C=10$ trained on binary MNIST. Each row contains 15 generated digits from a different cluster. Most clusters capture a single digit. Note that clusters have been ordered by hand so that generated digits appear in ascending order.

## 5.6 Variational RNN

### 5.6.1 Generative model

Imagine treating binary MNIST images as sequences: a 28-by-28 image of a digit can be thought of as a vertically stacked sequence of 28 length-28 vectors. Then, the generative model as specified by Variational Recurrent Neural Network (VRNN) [2] for such data is (for $t=1,2,\ldots,T=28$)

$$\begin{aligned} \boldsymbol{x}_t &\sim \text{ProductOfBernoullis}(\boldsymbol{p_\theta}(\boldsymbol{x}_{<t}, \boldsymbol{z}_{\leq t})) \\ \boldsymbol{z}_t &\sim \mathcal{N}(\boldsymbol{\mu_\theta}(\boldsymbol{x}_{<t}, \boldsymbol{z}_{<t}), \boldsymbol{\sigma_\theta}(\boldsymbol{x}_{<t}, \boldsymbol{z}_{<t})). \end{aligned}$$

where we have ignored the index $i$ for indexing training examples for brevity. $\boldsymbol{z}_t \in \mathbb{R}^L$ $(t=1,\ldots,T)$ are latent variables and $\boldsymbol{x}_t \in \{0,1\}^D$ $(t=1,\ldots,T)$ are observed variables. $\boldsymbol{\mu_\theta}$, $\boldsymbol{\sigma_\theta}$ and $\boldsymbol{p_\theta}$ are parametrized functions, which we will later define. Having latent and observed variables distributed over $T$ timesteps seems strange as compared to the simple formulation of latent variable model discussed in Section 1, but similar to what we did for GMVAE we could define the latent variable to be $\boldsymbol{z}=\boldsymbol{z}_1 \,\|\, \cdots \,\|\, \boldsymbol{z}_T$ with dimension $L'=T\times L$ and the observed variable to be $\boldsymbol{x}=\boldsymbol{x}_1 \,\|\, \cdots \,\|\, \boldsymbol{x}_T$ with dimension $D'=T\times D$, where $\|$ denotes concatenation. Notice how different latent variable models differ primarily in how the joint density $p_{\boldsymbol{\theta}}(\boldsymbol{x},\boldsymbol{z})$ factors. In fact, the structure of this model is inspired from the *complete* factorization of this joint density:

$$p(\boldsymbol{x},\boldsymbol{z}) = p(\boldsymbol{x}_{1:T}, \boldsymbol{z}_{1:T}) = p(\boldsymbol{z}_1)p(\boldsymbol{x}_1|\,\boldsymbol{z}_1)p(\boldsymbol{z}_2|\,\boldsymbol{x}_1,\boldsymbol{z}_1)p(\boldsymbol{x}_2|\,\boldsymbol{x}_1,\boldsymbol{z}_{1:2})\cdots,$$

which is different from Hidden Markov models, in which the latent variables are not conditioned on the observed variables, and observed variables are not conditioned on past observed variables.

We can rewrite the generative model more concisely using the following notation:

$$\begin{aligned} \boldsymbol{x}_t &\sim \text{ProductOfBernoullis}(\boldsymbol{p_\theta}(\boldsymbol{h}_t, \boldsymbol{z}_t)) \\ \boldsymbol{z}_t &\sim \mathcal{N}(\boldsymbol{\mu_\theta}(\boldsymbol{h}_t), \boldsymbol{\sigma_\theta}(\boldsymbol{h}_t)) \\ \boldsymbol{h}_t &= (\boldsymbol{x}_{<t}, \boldsymbol{z}_{<t}) \quad (\boldsymbol{h}_1 = ()), \end{aligned}$$

which motivates representing the model with a recurrent function $f$ with fixed-size $\boldsymbol{h}_t$:

$$\begin{aligned} \boldsymbol{x}_t &\sim \text{ProductOfBernoullis}(\boldsymbol{p_\theta}(\boldsymbol{h}_t, \boldsymbol{z}_t)) \\ \boldsymbol{z}_t &\sim \mathcal{N}(\boldsymbol{\mu_\theta}(\boldsymbol{h}_t), \boldsymbol{\sigma_\theta}(\boldsymbol{h}_t)) \\ \boldsymbol{h}_t &= f_{\boldsymbol{\theta}}(\boldsymbol{h}_{t-1}, \boldsymbol{x}_{t-1}, \boldsymbol{z}_{t-1}) \quad (\boldsymbol{h}_1 = \text{zeros}). \end{aligned}$$

where $f_{\boldsymbol{\theta}}: \mathbb{R}^H \times \{0,1\}^D \times \mathbb{R}^L \to \mathbb{R}^H$, $\boldsymbol{\mu}_{\boldsymbol{\theta}}: \mathbb{R}^H \to \mathbb{R}^L$, $\boldsymbol{\sigma}_{\boldsymbol{\theta}}: \mathbb{R}^H \to \mathbb{R}^{+L}$ and $\boldsymbol{p}_{\boldsymbol{\theta}}: \mathbb{R}^H \times \mathbb{R}^L \to \{0,1\}^D$ are neural networks. $\boldsymbol{\mu}_{\boldsymbol{\theta}}$, $\boldsymbol{\sigma}_{\boldsymbol{\theta}}$ and $\boldsymbol{p}_{\boldsymbol{\theta}}$ can be represented using feedforward neural networks, whereas $f$ needs to be represented using a recurrent neural network, such as an LSTM [6]. Note that an LSTM maintains a "cell state" (which is not processed by consecutive layers) in addition to the hidden state; with LSTM, the model is written as:

$$\begin{aligned}
\boldsymbol{x}_t &\sim \text{ProductOfBernoullis}(\boldsymbol{p}_{\boldsymbol{\theta}}(\boldsymbol{h}_t, \boldsymbol{z}_t)) \\
\boldsymbol{z}_t &\sim \mathcal{N}(\boldsymbol{\mu}_{\boldsymbol{\theta}}(\boldsymbol{h}_t), \boldsymbol{\sigma}_{\boldsymbol{\theta}}(\boldsymbol{h}_t)) \\
\underbrace{(\boldsymbol{h}_t, \boldsymbol{c}_t)}_{\text{state}_t} &= \text{LSTM}_{\boldsymbol{\theta}}\left(\boldsymbol{x}_{t-1}, \boldsymbol{z}_{t-1}, \underbrace{(\boldsymbol{h}_{t-1}, \boldsymbol{c}_{t-1})}_{\text{state}_{t-1}}\right).
\end{aligned}$$

### 5.6.2 Approximate posterior

As for GMVAE, to choose the form of approximate posterior, we first write out the factorization of the true posterior, which is intractable:

$$p_{\boldsymbol{\theta}}(\boldsymbol{z}_{1:T} | \boldsymbol{x}_{1:T}) = p_{\boldsymbol{\theta}}(\boldsymbol{z}_1 | \boldsymbol{x}_{1:T}) p_{\boldsymbol{\theta}}(\boldsymbol{z}_1 | \boldsymbol{z}_2, \boldsymbol{x}_{1:T}) p_{\boldsymbol{\theta}}(\boldsymbol{z}_3 | \boldsymbol{z}_{1:2}, \boldsymbol{x}_{1:T}) \cdots.$$

As in the original paper, we now make two simplications to the equation above to get the form of the approximate posterior. First, we use $q_{\boldsymbol{\theta}}(\boldsymbol{z}_t | \boldsymbol{x}_{\le t}, \boldsymbol{z}_{<t})$ to approximate each $p_{\boldsymbol{\theta}}(\boldsymbol{z}_t | \boldsymbol{x}_{1:T}, \boldsymbol{z}_{<t})$, which ignores the dependency of $\boldsymbol{z}_t$ on $\boldsymbol{x}_{>t}$: since $\boldsymbol{z}_t$ contributes to all future $\boldsymbol{x}_{>t}$ in the generative procedure, knowing the values of $\boldsymbol{x}_{>t}$ should inform us probabilistically about the value of $\boldsymbol{z}_t$. The advantage of doing so is that now $q_{\boldsymbol{\theta}}(\boldsymbol{z}_t | \boldsymbol{x}_{\le t}, \boldsymbol{z}_{<t})$ can be written as $q_{\boldsymbol{\theta}}(\boldsymbol{z}_t | \boldsymbol{x}_t, \boldsymbol{h}_t)$, where $\boldsymbol{h}_t$ also appeared in the generative model, and hence the generative and inference networks can share the same recurrent neural network. An interested reader should implement $q_{\boldsymbol{\theta}}(\boldsymbol{z}_t | \boldsymbol{x}_{1:T}, \boldsymbol{z}_{<t})$ as a practice. Secondly, we choose $q_{\boldsymbol{\theta}}(\boldsymbol{z}_t | \boldsymbol{x}_{\le t}, \boldsymbol{z}_{<t})$ to be a diagonal Gaussian, as in VAE.

### 5.6.3 Estimator for per-example ELBO

Following the approach in Section 5.2.3, we first derive an unbiased, low-variance estimator of the per-example ELBO whose randomness does not depend on parameters $\boldsymbol{\theta}$, $\boldsymbol{\phi}$. Starting from the per-example ELBO, we have (following the derivation in Appendix of [2]; color highlights the trick used in the derivation):

$$\begin{aligned}
&\mathbb{E}_{q_{\boldsymbol{\phi}}(\boldsymbol{z}_{1:T} | \boldsymbol{x}_{1:T})}[\log p_{\boldsymbol{\theta}}(\boldsymbol{x}_{1:T}, \boldsymbol{z}_{1:T}) - \log q_{\boldsymbol{\phi}}(\boldsymbol{z}_{1:T} | \boldsymbol{x}_{1:T})] \\
=\; &\mathbb{E}_{q_{\boldsymbol{\phi}}(\boldsymbol{z}_{1:T} | \boldsymbol{x}_{1:T})}\left[\log \prod_{t=1}^{T} p_{\boldsymbol{\theta}}(\boldsymbol{x}_t | \boldsymbol{x}_{<t}, \boldsymbol{z}_{\le t}) p_{\boldsymbol{\theta}}(\boldsymbol{z}_t | \boldsymbol{x}_{<t}, \boldsymbol{z}_{<t}) - \log \prod_{t=1}^{T} q_{\boldsymbol{\phi}}(\boldsymbol{z}_t | \boldsymbol{x}_{\le t}, \boldsymbol{z}_{<t})\right] \\
=\; &\mathbb{E}_{q_{\boldsymbol{\phi}}(\boldsymbol{z}_{1:T} | \boldsymbol{x}_{1:T})}\left[\sum_{t=1}^{T} \log p_{\boldsymbol{\theta}}(\boldsymbol{x}_t | \boldsymbol{x}_{<t}, \boldsymbol{z}_{\le t}) + \log p_{\boldsymbol{\theta}}(\boldsymbol{z}_t | \boldsymbol{x}_{<t}, \boldsymbol{z}_{<t}) - \log q_{\boldsymbol{\phi}}(\boldsymbol{z}_t | \boldsymbol{x}_{\le t}, \boldsymbol{z}_{<t})\right] \\
=\; &\sum_{t=1}^{T} \mathbb{E}_{q_{\boldsymbol{\phi}}(\boldsymbol{z}_{\le t} | \boldsymbol{x}_{\le t})}[\log p_{\boldsymbol{\theta}}(\boldsymbol{x}_t | \boldsymbol{x}_{<t}, \boldsymbol{z}_{\le t})] - \mathbb{E}_{q_{\boldsymbol{\phi}}(\boldsymbol{z}_{\le t} | \boldsymbol{x}_{\le t})}[\log q_{\boldsymbol{\phi}}(\boldsymbol{z}_t | \boldsymbol{x}_{\le t}, \boldsymbol{z}_{<t}) - \log p_{\boldsymbol{\theta}}(\boldsymbol{z}_t | \boldsymbol{x}_{<t}, \boldsymbol{z}_{<t})] \\
=\; &\sum_{t=1}^{T} \mathbb{E}_{\color{red}{q_{\boldsymbol{\phi}}(\boldsymbol{z}_{\le t} | \boldsymbol{x}_{\le t})}}[\log p_{\boldsymbol{\theta}}(\boldsymbol{x}_t | \boldsymbol{x}_{<t}, \boldsymbol{z}_{\le t})] - \mathbb{E}_{\color{red}{q_{\boldsymbol{\phi}}(\boldsymbol{z}_{<t} | \boldsymbol{x}_{<t})}}[D_{\mathbb{KL}}(q_{\boldsymbol{\phi}}(\boldsymbol{z}_t | \boldsymbol{x}_{\le t}, \boldsymbol{z}_{<t}) \,\|\, p_{\boldsymbol{\theta}}(\boldsymbol{z}_t | \boldsymbol{x}_{<t}, \boldsymbol{z}_{<t}))] \\
=\; &\sum_{t=1}^{T} \mathbb{E}_{\color{red}{q_{\boldsymbol{\phi}}(\boldsymbol{z}_{1:T} | \boldsymbol{x}_{1:T})}}[\log p_{\boldsymbol{\theta}}(\boldsymbol{x}_t | \boldsymbol{x}_{<t}, \boldsymbol{z}_{\le t})] - \mathbb{E}_{\color{red}{q_{\boldsymbol{\phi}}(\boldsymbol{z}_{1:T} | \boldsymbol{x}_{1:T})}}[D_{\mathbb{KL}}(q_{\boldsymbol{\phi}}(\boldsymbol{z}_t | \boldsymbol{x}_{\le t}, \boldsymbol{z}_{<t}) \,\|\, p_{\boldsymbol{\theta}}(\boldsymbol{z}_t | \boldsymbol{x}_{<t}, \boldsymbol{z}_{<t}))] \\
=\; &\mathbb{E}_{q_{\boldsymbol{\phi}}(\boldsymbol{z}_{1:T} | \boldsymbol{x}_{1:T})}\left[\sum_{t=1}^{T} \log p_{\boldsymbol{\theta}}(\boldsymbol{x}_t | \boldsymbol{x}_{<t}, \boldsymbol{z}_{\le t}) - D_{\mathbb{KL}}(q_{\boldsymbol{\phi}}(\boldsymbol{z}_t | \boldsymbol{x}_{\le t}, \boldsymbol{z}_{<t}) \,\|\, p_{\boldsymbol{\theta}}(\boldsymbol{z}_t | \boldsymbol{x}_{<t}, \boldsymbol{z}_{<t}))\right],
\end{aligned}$$

where we can remove $\boldsymbol{\phi}$ from under the expectation by first decomposing $q_{\boldsymbol{\phi}}(\boldsymbol{z}_{1:T}|\, \boldsymbol{x}_{1:T})$ into $q_{\boldsymbol{\phi}}(\boldsymbol{z}_1|\, \boldsymbol{x}_1)\cdots q_{\boldsymbol{\phi}}(\boldsymbol{z}_T|\, \boldsymbol{x}_{\leq T}, \boldsymbol{z}_{<T})$ and then applying reparametrization trick to each component distribution (since each component distribution is a Gaussian):

$$\mathbb{E}_{\boldsymbol{\varepsilon}_1 \sim \mathcal{N}(0,\boldsymbol{I}_L),\ldots,\boldsymbol{\varepsilon}_T \sim \mathcal{N}(0,\boldsymbol{I}_L)} \left[ \sum_{t=1}^{T} \log p_{\boldsymbol{\theta}}(\boldsymbol{x}_t|\, \boldsymbol{x}_{<t}, \boldsymbol{z}^s_{\leq t}) - D_{\mathbb{KL}}(q_{\boldsymbol{\phi}}(\boldsymbol{z}_t|\, \boldsymbol{x}_{\leq t}, \boldsymbol{z}_{<t}) \,\|\, p_{\boldsymbol{\theta}}(\boldsymbol{z}_t|\, \boldsymbol{x}_{<t}, \boldsymbol{z}_{<t})) \right],$$

where $\boldsymbol{z}^s_{1:T}$ is computed from a deterministic transformation of $\boldsymbol{\varepsilon}_{1:T}$ and $\boldsymbol{\theta}$:

$$\begin{aligned} \boldsymbol{z}_1 &= \boldsymbol{\mu}_{\boldsymbol{\theta}}(\boldsymbol{x}_1) + \boldsymbol{\sigma}_{\boldsymbol{\theta}}(\boldsymbol{x}_1)\boldsymbol{\varepsilon}_1 \\ \boldsymbol{z}_2 &= \boldsymbol{\mu}_{\boldsymbol{\theta}}(\boldsymbol{x}_{\leq 2}, \boldsymbol{z}_1) + \boldsymbol{\sigma}_{\boldsymbol{\theta}}(\boldsymbol{x}_{\leq 2}, \boldsymbol{z}_1)\boldsymbol{\varepsilon}_2 \\ &\vdots \end{aligned}$$

which can also be written in terms of hidden states of the LSTM:

$$\begin{aligned} \boldsymbol{h}_1, \boldsymbol{c}_1 &= (\text{zeros}, \text{zeros}) \\ \boldsymbol{z}_1 &= \boldsymbol{\mu}_{\boldsymbol{\theta}}(\boldsymbol{x}_1, \boldsymbol{h}_1) + \boldsymbol{\sigma}_{\boldsymbol{\theta}}(\boldsymbol{x}_1, \boldsymbol{h}_1)\boldsymbol{\varepsilon}_1 \\ \boldsymbol{h}_2, \boldsymbol{c}_2 &= \text{LSTM}_{\boldsymbol{\theta}}(\boldsymbol{x}_1, \boldsymbol{z}_1, (\boldsymbol{h}_1, \boldsymbol{c}_1)) \\ \boldsymbol{z}_2 &= \boldsymbol{\mu}_{\boldsymbol{\theta}}(\boldsymbol{x}_2, \boldsymbol{h}_2) + \boldsymbol{\sigma}_{\boldsymbol{\theta}}(\boldsymbol{x}_2, \boldsymbol{h}_2)\boldsymbol{\varepsilon}_2 \\ &\vdots \end{aligned}$$

We can now instantiate an unbiased estimator:

$$\sum_{t=1}^{T} \log p(\boldsymbol{x}_t|\, \boldsymbol{x}_{<t}, \boldsymbol{z}^s_{\leq t}) - D_{\mathbb{KL}}(q_{\boldsymbol{\phi}}(\boldsymbol{z}_t|\, \boldsymbol{x}_{\leq t}, \boldsymbol{z}_{<t}) \,\|\, p(\boldsymbol{z}_t|\, \boldsymbol{x}_{<t}, \boldsymbol{z}_{<t})).$$

which can also be written in terms of hidden states of the LSTM as follows:

$$\sum_{t=1}^{T} \log p(\boldsymbol{x}_t|\, \boldsymbol{z}^s_t, \boldsymbol{h}_t) - D_{\mathbb{KL}}(q_{\boldsymbol{\phi}}(\boldsymbol{z}_t|\, \boldsymbol{x}_t, \boldsymbol{h}_t) \,\|\, p(\boldsymbol{z}_t|\, \boldsymbol{h}_t)),$$

where the $\boldsymbol{h}_t$'s here are exactly the $\boldsymbol{h}_t$'s used in the process of sampling $\boldsymbol{z}^s_{1:T}$. In PyTorch, we can implement this estimator and compute the gradients with respect to the parameters as follows:

```
class AEVB(nn.Module):
    # ...
    def step(self, x):
        # x shape (batch_size, 28, 28)

        bs = x.size(0)
        seq_len = x.size(1)

        mini_batch_elbo = 0  # mean of estimators of per-example ELBO

        for t in range(seq_len):

            xt = x[:,t,:]

            if t == 0:
                ht = torch.zeros(bs, self.h_dim)
                state_t = None
            else:
                ht, state_t = self.f(x_tminus1, z_tminus1, state_tminus1)
                ht = ht.squeeze()  # get rid of the "direction" dimension

            prior_over_zt = self.p_zt_given_ht(ht)
            posterior_over_zt = self.q_zt_given_ht_xt(ht, xt)
```

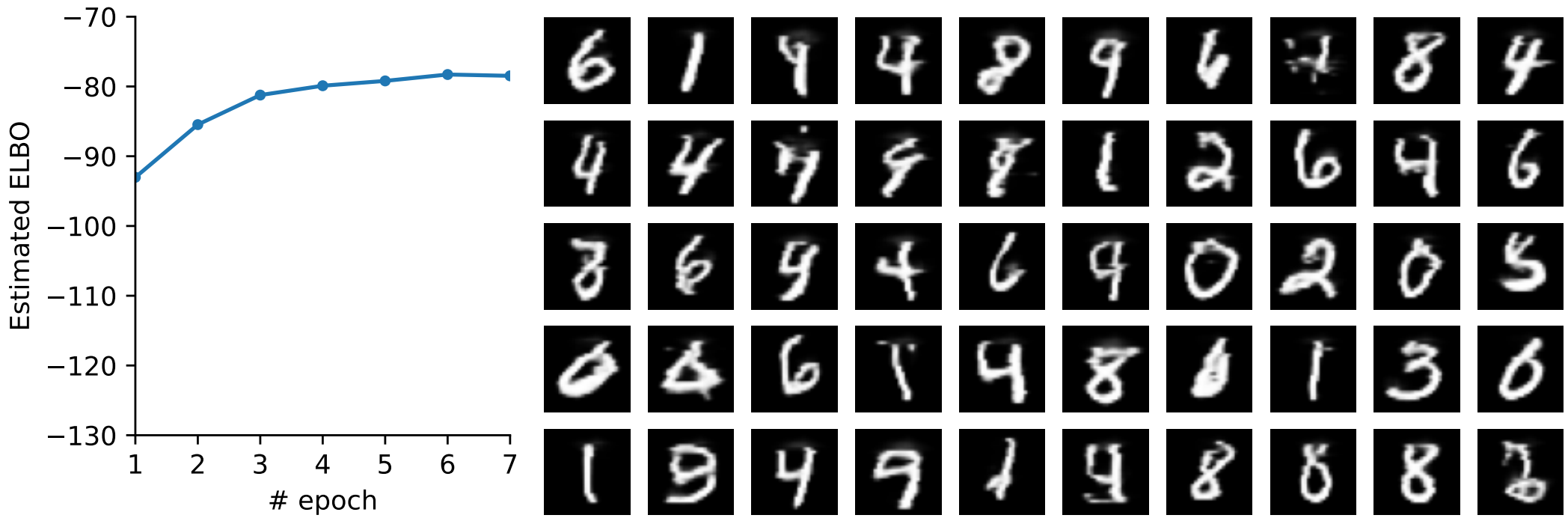

**Figure 12.** Results for a VRNN with $L = 2$ trained on binary MNIST sequences. (left) Estimated ELBO on the test set over time (right) Generated images after training. Each image is generated one row at a time.

```
        kl = kl_divergence(posterior_over_zt, prior_over_zt)
        zt = posterior_over_zt.sample()
        rec = self.p_xt_given_ht_zt(ht, zt).log_prob(xt)  # reconstruction

        mini_batch_elbo += (rec - kl).mean()

        x_tminus1 = xt
        z_tminus1 = zt
        state_tminus1 = state_t

    loss = - mini_batch_elbo

    self.opt.zero_grad()
    loss.backward()
    self.opt.step()
# ...
```

### 5.6.4 Results

We transformed each MNIST image into a length-28 sequence of rows, each of which is a length-28 vector. We then trained a VRNN on these sequences by optimizing a mini-batch estimator of ELBO via gradient ascent until convergence. We used a small latent dimension $L = 2$ since $T = 28$, a mini-batch size of 100 and Adam with a learning rate of 0.001. Figure 12 (left) shows that the estimated ELBO on the test set improves over time, and we applied early stopping when the quality of generations became satisfactory. Figure 11 (right) shows images generated by VRNN after training; images were generated one row at a time.